\documentclass{article} % For LaTeX2e
\usepackage{main,times}

\usepackage{amsmath,amsfonts,bm}

\def\eqref#1{equation~\ref{#1}}
\def\1{\bm{1}}

\def\mA{{\bm{A}}}
\def\mB{{\bm{B}}}

\def\mH{{\bm{H}}}

\def\mW{{\bm{W}}}
\def\mX{{\bm{X}}}

\DeclareMathAlphabet{\mathsfit}{\encodingdefault}{\sfdefault}{m}{sl}
\SetMathAlphabet{\mathsfit}{bold}{\encodingdefault}{\sfdefault}{bx}{n}

\def\sM{{\mathbb{M}}}

\def\sT{{\mathbb{T}}}

\definecolor{CColor}{rgb}{0.01,0.31,0.59}
\definecolor{GGray}{rgb}{0.80,0.90,1}
\definecolor{Shady}{rgb}{0.9,0.9,0.9}
\definecolor{kaistblue}{RGB}{20,135,200}
\definecolor{kaistdarkblue}{RGB}{0,65,145}
\definecolor{urbanablue}{RGB}{19,41,75}
\definecolor{urbanaorange}{RGB}{232,74,39}
\definecolor{drp}{rgb}{0.53,0.15,0.34}
\usepackage[plainpages=false,pdfpagelabels,colorlinks=true,linkcolor=CColor,citecolor=CColor,urlcolor=CColor]{hyperref}

\usepackage{url}
\usepackage{graphicx} 
\usepackage{subcaption}
\usepackage{multirow}
\usepackage{enumitem}
\usepackage{algorithm}
\usepackage{algorithmic}
\title{Solving Continual Offline Reinforcement Learning with Decision Transformer}
\iclrfinalcopy
\author{
Kaixin Huang\textsuperscript{\rm 1}
\quad 
Li Shen\textsuperscript{\rm 2}\thanks{Corresponding authors} 
\quad 
Chen Zhao\textsuperscript{\rm 1}
\quad 
Chun Yuan\textsuperscript{\rm 1}\textsuperscript{\textasteriskcentered}
\quad 
Dacheng Tao\textsuperscript{\rm 3}
\\
\textsuperscript{\rm 1}Tsinghua Shenzhen International Graduate School, China\\
\textsuperscript{\rm 2}JD Explore Academy, China \\
\textsuperscript{\rm 3}The University of Sydney, Australia\\
{\tt\small \{huangkx21,zhao-c21\}@mails.tsinghua.edu.cn; mathshenli@gmail.com}\\
{\tt\small yuanc@sz.tsinghua.edu.cn; dacheng.tao@gmail.com
}
}

\begin{document}

\maketitle

% \begin{abstract}
% We delve into the realm of continual offline reinforcement learning (CORL), a practical paradigm in the field of reinforcement learning that learns a sequence of reinforcement learning tasks from offline datasets without forgetting previous tasks. The previous methods mainly combined the offline algorithm based on the actor-critic structure and experience replay which has many problems such as multiple distribution shifts, high dependence on the quality of the replay buffer and weak knowledge-sharing capabilities. In this paper, we explore and conclude that Decision Transformer (DT) is a more suitable offline learning algorithm for the CORL problem and focus on solving the more serious forgetting problem of DT. We propose MH-DT \ls{why this name, it is not clear}, a method based on common structure and selective experience replay, and LoRA-DT, a method based on weight fusion and low-rank adaptation, to solve the CORL problem when replay buffers are available and unavailable respectively. Extensive experimentation demonstrates that our methods outperform SOTA baselines in all CORL metrics and show additional advantages such as robustness when the number of tasks increases and memory efficiency compared to saving buffers. We believe that DT with continuous learning capabilities and avoidance of forgetting can help the development of large models for decision-making. \ls{reorganize.}
% \end{abstract}

\begin{abstract}

Continuous offline reinforcement learning (CORL) combines continuous and offline reinforcement learning, enabling agents to learn multiple tasks from static datasets without forgetting prior tasks. However, CORL faces challenges in balancing stability and plasticity. Existing methods, employing Actor-Critic structures and experience replay (ER), suffer from distribution shifts, low efficiency, and weak knowledge-sharing. We aim to investigate whether Decision Transformer (DT), another offline RL paradigm, can serve as a more suitable offline continuous learner to address these issues. We first compare AC-based offline algorithms with DT in the CORL framework. DT offers advantages in learning efficiency, distribution shift mitigation, and zero-shot generalization but exacerbates the forgetting problem during supervised parameter updates.
We introduce multi-head DT (MH-DT) and low-rank adaptation DT (LoRA-DT) to mitigate DT's forgetting problem. MH-DT stores task-specific knowledge using multiple heads, facilitating knowledge sharing with common components. It employs distillation and selective rehearsal to enhance current task learning when a replay buffer is available. In buffer-unavailable scenarios, LoRA-DT merges less influential weights and fine-tunes DT's decisive MLP layer to adapt to the current task.
Extensive experiments on MoJuCo and Meta-World benchmarks demonstrate that our methods outperform SOTA CORL baselines and showcase enhanced learning capabilities and superior memory efficiency.
\end{abstract}

\section{Introduction}
% Unlike machines, humans possess a remarkable capacity for continuous learning and lifelong adaptation, a phenomenon known as continuous learning (CL). Previous research has primarily focused on endowing reinforcement learning algorithms with the ability to continuously learn, enabling them to rapidly acquire new tasks in sequential interactions with online environments while retaining knowledge of past tasks ~\citep{rolnick_ahuja_schwarz_lillicrap_wayne_2018,isele_cosgun_2022,zenke_poole_ganguli_2017}.
% However, in certain scenarios, acquiring new data and engaging with the environment may incur high costs or result in severe consequences. Hence, there is significant value in leveraging existing data to facilitate efficient, safe, and transferable learning. 
Continuous offline reinforcement learning (CORL) \citep{gai2023offline} is an innovative paradigm that merges the principles of continuous learning with offline reinforcement learning. CORL aims to empower agents to learn multiple tasks from static offline datasets and swiftly adapt to new, unknown tasks.
A central and persistent challenge in CORL is the delicate balance between plasticity and stability  \citep{khetarpal2022towards}. On one hand, the reinforcement learning policy must preserve knowledge and prevent forgetting of historical tasks (stability). On the other hand, it should exhibit the ability to rapidly adapt to new tasks (plasticity). 
The best way to achieve this balance is to retain common knowledge between tasks during the continuous learning process (stability) and to retain only the most important common knowledge as much as possible to prevent the loss of plasticity. It is worth noting that selecting the common knowledge between tasks in RL problems is non-trivial because unlike in CV tasks where features between images are explicit common knowledge, it is difficult to infer feature knowledge from offline trajectories that can be universal across tasks.

Existing methods predominantly integrate offline algorithms based on actor-critic structures with continual learning techniques ~\citep{gai2023offline}, with rehearsal-based methods ~\citep{lopez2017gradient,chaudhry2019efficient} being the most commonly employed. However, these methods encounter several challenges, including multiple distribution shifts, suboptimal learning efficiency, and limited knowledge-sharing capabilities.
These distribution shifts manifest in three forms.
Firstly, there is a distribution shift between the behavior policy and the learning policy inherent to AC-based offline algorithms. 
% an inherent characteristic of all offline algorithms employing Actor-Critic structures. This shift inevitably gives rise to the overestimation problem with out-of-distribution (OOD) data. 
Secondly, distribution shifts occur between the offline data from different tasks, leading to catastrophic forgetting. 
Lastly, distribution shifts between the learned policy and the saved replay buffer which result in performance degradation in previous tasks during the rehearsal process.
% The low learning efficiency can be attributed to the AC-based offline algorithm's simultaneous learning of the Q function and policy, making it challenging to distil the knowledge from the learned policy into the current agent within limited training steps.
Regarding knowledge-sharing capabilities, while the relevance of the Q function in related tasks has been established ~\citep{van2019composing}, these methods merely introduce behavioral clones to the actor which shares few knowledge since the actor's objective is to maximize the Q value.

Decision Transformer (DT) \citep{chen2021decision}, another offline RL paradigm, shows extremely strong learning efficiency and can ignore the problem of distribution shift in offline RL because its supervised learning training method. In addition, previous work has proposed to use the power of architecture inductive bias of the transformer model to achieve zero-shot generalization\citep{xu2022prompting}, indicating that DT trained through a large amount of data can have multi-task capabilities. Inspired by DT’s outstanding characteristics and previous multi-task work, we aim to investigate whether DT can serve as a more suitable offline continuous learner in this work.

% To solve the above problems, we first rethink the process of  CORL by comparing an AC-based offline RL algorithm with Decision Transformer (DT) \cite{chen2021decision}, another offline RL paradigm. 
We first rethink the process of  CORL by comparing an AC-based offline RL algorithm with Decision Transformer (DT).
The results underscore several advantages of DT,  including heightened learning efficiency, bypassing distribution shift of offline learning, and superior zero-shot generalization capabilities. However, the forgetting issue of DT is more serious, manifested by a rapid decline in performance after switching tasks. This heightened sensitivity to distribution shifts between time-evolving datasets is attributed to DT's training using supervised learning and updating all parameters.

In order to retain the advantages of DT and solve the more serious problem of catastrophic forgetting, we propose multi-head DT (MH-DT) and low-rank adaptation DT (LoRA-DT). MH-DT uses multiple heads to store task-specific knowledge and share knowledge with common parts, to avoid the catastrophic forgetting problem caused by all parameters being changed when the dataset distribution shift occurs. Besides, using the structural characteristics of the transformer and the feature that DT will benefit from training on close tasks we propose an additional distillation objective and selective rehearsal module to improve the learning effect of the current task.
To solve the CORL problem when the replay buffer is not allowed – for example, in real-world scenarios where data privacy matters \citep{shokri2015privacy}, inspired by \cite{lawson2023merging} that explored the similarities and differences of each module in the DT structure under multi-task situations, we proposed LoRA-DT that merges weights that have
little impact for knowledge sharing and fine-tunes the decisive MLP layer in DT blocks with LoRA to adapt to the current task. This method also avoids substantial performance deterioration with a smaller buffer size \citep{cha2021co2l}.

Extensive experiments on MuJoCo \citep{todorov2012mujoco} and Meta-World \citep{yu2020meta} benchmarks demonstrate that our methods outperform SOTA baselines in all CORL metrics. Our DT-based methods also show other advantages including stronger learning ability and more memory-efficient. We also experimentally prove that simply combining DT with the continuous learning method in supervised learning cannot solve the CORL problem well.
% For example, the learning effect of MH-DT is more robust when the number of review tasks increases. 
% Saving the updated $\mA\mB$ matrix to avoid forgetting previous tasks in LoRA-DT has also proven to be a more memory-efficient method compared with saving replay buffers.
% In summary, this paper rethinks the CORL problem with decision transformer and points out the advantages of using DT to solve CORL and the problems that need to be solved. To leverage the benefits and solve the problem of catastrophic forgetting, we propose two methods, MH-DT and LORA-DT for the scenarios where the replay buffer is available and unavailable respectively. 
The main contributions of this paper can be summarized as four folds:
\begin{itemize}[]
    \item We propose that compared to the offline RL methods employing the Actor-Critic structure, methods utilizing DT as the foundational model are better suited for addressing the CORL problem and point out the advantages of DT and problems that need to be solved by rethinking the CORL process with decision transformer. To the best of our knowledge, we are the first to propose using DT as an underlying infrastructure in CORL setting. 
    \item When the replay buffer is available, we propose MH-DT to solve the problem of catastrophic forgetting and avoid the problem of low learning efficiency by distillation and selective rehearsal.
    \item We propose LoRA-DT that uses weights merging and low-rank adaptation to avoid catastrophic forgetting and save memory by saving the fine-tuned updated matrix when the buffer is unavailable.
    \item We experimentally demonstrate that our methods outperform not only prior CORL methods but also simple combinations of DT and continual learning methods, perform better learning capability, and are more memory-efficient.
\end{itemize}

\section{Related Work}

% \subsection{Offline reinforcement learning}
\textbf{Offline reinforcement learning.}
Offline reinforcement learning allows policy learning from data collected by arbitrary policies, increasing the sample efficiency of RL. The key issues in offline reinforcement learning are distributional shift and value overestimation. Some prior model-free works suggest constraining the learned policy to align with the behavior policy using regularization methods\citep{peng2019advantageweighted, nair2021awac, wang2020critic,dadashi2021offline,kumar2019stabilizing}.
Alternatively, some model-based approaches\citep{yu2020mopo,yu2021combo} advocate firstly training a dynamic model in a supervised learning way and then mitigating the impact of out-of-distribution (OOD) sample by the generalizability of the model itself or adding a constrain estimated by model uncertainty.
% to predict out-of-distribution (OOD) sample values via supervised learning techniques.Other works \cite{yu2020mopo,yu2021combo} propose to train a dynamic model to predict the values of OOD samples in a supervised learning way. 
% Model-based methods are believed to mitigate the impact of OOD and thereby improve the robustness of offline agents.
% Such model-based offline RL methods offer great potential for solving the OOD problem, even though the transition model is hardly accurate strictly. Model-based methods are believed to mitigate the impact of OOD and thereby improve the robustness of offline agents. 
However, all these methods are based on AC structure, and their final performance depends on the accuracy of Q-value estimation. 
% Although previous methods use multiple methods to reduce the harm of estimation errors. 
These methods cannot achieve good results in offline continuous reinforcement learning since it's difficult to obtain accurate estimates when distribution shifts and there is no obvious relationship between the actor-network of different tasks.
Recently, \cite{chen2021decision} proposed Decision Transformer(DT) to solve offline reinforcement learning problems by casting RL problem as conditional sequence modelling.
DT bypasses the need for bootstrapping for long term credit assignment – thereby avoiding one of the "deadly triad"\citep{sutton2018reinforcement} known to destabilize RL.
DT demonstrates superior learning efficiency compared to AC-structured algorithms. 
Compared with simple imitation learning, DT also has the ability to combine offline trajectories based on the attention mechanism. 
In this paper, we propose that DT is more suitable for offline continuous reinforcement learning scenarios, use DT as a backbone network and focus on solving its catastrophic forgetting problem.

% \subsection{Continual Reinforcement Learning}
\textbf{Continual Reinforcement Learning.}
Continual learning is a challenging and important problem in machine learning, where the goal is to enable a model to learn from a stream of tasks without forgetting the previous ones. Generally, continual learning methods can be classified into three categories \citep{parisi2019continual}. Regularization-based methods usually balance old and new tasks by adding explicit regularization terms in the loss function\citep{kirkpatrick2017overcoming,zenke2017continual,li2017learning}. Replay-based methods attempt to approximate or recover the distribution of old data\citep{chaudhry2019tiny,riemer2019learning,shin2017continual,wu2018memory}. Modular approaches construct task-specific parameters instead of learning all incremental tasks with a shared set of parameters\citep{mallya2018piggyback,mallya2018packnet,serra2018overcoming,ebrahimi2020adversarial}.
% regularization-based approaches \cite{kirkpatrick2017overcoming,zenke2017continual} add a regularization term to prevent the parameters from far from the value learned from past tasks; modular approaches \cite{fernando2017pathnet,mallya2018packnet} consider fixed partial parameters for a dedicated task; and rehearsal-based methods \cite{chaudhry2019efficient,lopez2017gradient} train an agent by merging the data of previously learned tasks with that of the current task. 

Continual reinforcement learning is a more complex problem because there is no way to explicitly model the relationship between the task and the loss function. Many continual learning methods in supervised learning cannot be used in the decision-making tasks of reinforcement learning. Continual reinforcement learning methods can be divided into three categories\citep{khetarpal2022towards}: explicit knowledge retention, leverage shared structures, and learning to learn\citep{ammar2014online,borsa2016learning,yu2020gradient,shi2021meta}.

These methods mainly considered online reinforcement learning. In CORL problem setting, \cite{gai2023offline} demonstrated that Experience Replay (ER), a rehearsal-based method, is the most effective. However, ER methods need to consider the stability-plasticity trade-off.
% , which means the RL policy on the one hand needs to retain and reuse the knowledge shared across different tasks in history (stability) while on the other hand can be quickly adapted to new tasks without interference from previous tasks (plasticity). 
The previous ER method was mainly based on AC structure and focused more on stability. As a result, there are too many tasks that need to be reviewed later in the learning process, resulting in failure to achieve good results in subsequent tasks. In this paper, we propose MH-DT, a method that applies a modular approach and experience replay method to DT, plus distillation objective and selective rehearsal for better plasticity. Then, considering the characteristics of DT in handling multi-tasks, we propose a new DT-based continuous learning method referred to as LoRA-DT, that uses weight merging to share knowledge and saves LoRA matrices to avoid forgetting.

% In this paper, We use DT's ability to perceive tasks and the weight similarity of similar tasks to propose selective review and achieve balance. And with the help of the special properties of transformer, we propose a method to use LoRA to save more space and avoid the review process. \ls{move this paragraph to other place.}

\textbf{Multi-task Decision Transformer.}
Many previous works have confirmed that DT can achieve multi-task capabilities or few-shot generalization to new tasks by training on offline multi-task datasets. \cite{xu2022prompting} 
propose Prompt-DT, which uses a short trajectory as a prompt to enable the DT agent to quickly adapt to unseen tasks. \cite{xu2023hyperdecision} propose Hyper-Decision Transformer, in which through a hypernetwork, the parameters of the adapter in the DT block are generated to achieve parameter-efficiency generalization to unseen tasks. \cite{lawson2023merging} propose a method for creating multi-task DTs through merging certain parameters, freezing those merged, and then independently finetuning un-merged parts. Recently, \citet{schmied2024learning} propose Learning-to-Modulate (L2M) to avoid the forgetting problem when fine-tuning the pre-trained model to adapt to new tasks.
These works are implemented in multi-task settings, that is, the offline data of multiple tasks can be accessed simultaneously. However, in practical scenarios, we may not be able to obtain data from multiple tasks at the same time due to issues like data privacy. Therefore, we want to explore whether the DT structure can learn multiple tasks in sequence from scratch and become an agent with multi-task capabilities. Since all data can be accessed simultaneously, multi-task methods are regarded as an upper bound for continuous learning methods.
\section{Rethinking CORL with Decision Transformer}

Below, we first review CORL problem setting and Decision Transformer. Then, we rethink the process of CORL using DT and compare it with offline algorithms using AC structure.

\subsection{Preliminary}\label{chp:pre}
\textbf{Continual Offline Reinforcement Learning } In this paper, we investigate CORL, which learns a sequence of RL tasks $ \sT = \{T_1, \cdots, T_N\}$. Each task $T_n$ is selected from a group of similar Markov Decision Process (MDP) family and can be represented by a tuple of $\left\{\mathcal{S}, \mathcal{A}, P_n, \rho_{0, n}, r_n, \gamma\right\}$, where $\mathcal{S}$ is the state space, $\mathcal{A}$ is the action space, $P_n: \mathcal{S} \times \mathcal{A} \times \mathcal{S} \leftarrow[0,1]$ is the transition probability, $\rho_{0, n}: \mathcal{S}$ is the distribution of the initial state, $r_n: \mathcal{S} \times \mathcal{A} \leftarrow\left[-R_{\max }, R_{\max }\right]$ is the reward function, and $\gamma \in(0,1]$ is the discounting factor. In this work, we only consider sequential tasks that have different $P_n, \rho_{0, n}$ and $r_n$, but share the same $\mathcal{S}, \mathcal{A}$, and $\gamma$. The return is defined as the sum of discounted future reward $R_{t, n}=\sum_{i=t}^H \gamma^{(i-t)} r_n\left(s_i, a_i\right)$, where $H$ is the horizon. $\mathcal{D}_n=\left\{e_n^i\right\}, e_n^i=\left(s_n^i, a_n^i, s_n^{\prime i}, r_n^i\right)$ are offline datasets of different tasks obtained by unknown policies $\pi_n^\beta(a \mid s)$.

\textbf{Decision Transformer (DT)} for offline RL treats learning a policy as a sequential modeling problem, which trains a transformer model to predict action according to a trajectory sequence $\tau$ autoregressively as input which contains the most recent $K$-step history $\tau=\left(\hat{r}_{t-K+1}, s_{t-K+1}, a_{t-K+1}, \ldots, \hat{r}_t, s_t, a_t\right)$. $\hat{r}_t$ is reward-to-go which is the cumulative rewards from the current step till the end of the episode that can help select actions with the most rewards. In the test phase, $\hat{r}_t=G^{\star}-\sum_{i=0}^t r_i$ is the target reward minus the accumulated rewards currently obtained. All $s_t, a_t$ and $\hat{r}_t$ are embedded into the same dimension and concatenated with timestep embeddings to serve as tokens in Transformer model.

% It proposes to model trajectories with state $s_t$, action $a_t$ and reward-to-go $\hat{r}_t$ tuples collected at different time steps $t$. The reward-to-go is the cumulative rewards from the current time step till the end of the episode. Instead of including the one-step reward $r_t$, this novel representation helps guide action selection towards optimizing the return. At timestep $t$, Decision Transformer takes a trajectory sequence $\tau$ autoregressively as input which contains the most recent $K$-step history $\tau=\left(\hat{r}_{t-K+1}, s_{t-K+1}, a_{t-K+1}, \ldots, \hat{r}_t, s_t, a_t\right)$.
% % \begin{equation}
% % \tau=\left(\hat{r}_{t-K+1}, s_{t-K+1}, a_{t-K+1}, \ldots, \hat{r}_t, s_t, a_t\right)
% % \end{equation}
% When training with offline collected data, $\hat{r}_t=\sum_{i=t}^T r_i$. During testing, $\hat{r}_t=G^{\star}-\sum_{i=0}^t r_i$ where $G^{\star}$ is the targeted total return for an episode. Each trajectory $\tau$ corresponds to $3 K$ tokens in the standard Transformer model. To encode the sequence timestep information, DT concatenates the same timestep embedding to the embeddings of $s_t, a_t$ and $\hat{r}_t$.
% % Each head corresponding to a state token 
% DT is trained to predict an action by minimizing mean-squared loss.
% when continuous action spaces.

% In this paper, we take advantage of the characteristics of DT and propose two DT-based methods, MH-DT and LORA-DT, to solve the CORL problem when the replay buffer is available and unavailable respectively.

\subsection{Rethinking CORL with Decision Transformer} \label{chp:rethink}
We conduct experiments in the Cheetal-Vel environment, training Vanilla DT alongside an offline algorithm TD3+BC  \citep{fujimoto2021minimalist} based on the AC structure.
We use six tasks and train 30K steps for each task as in Fig.\ref{fig:explore}. We hope to explore the following two questions: 
\begin{figure}[!h]
  \centering
  \subcaptionbox{Decision Transformer\label{fig:DT}}
    {\includegraphics[width=0.3\linewidth]{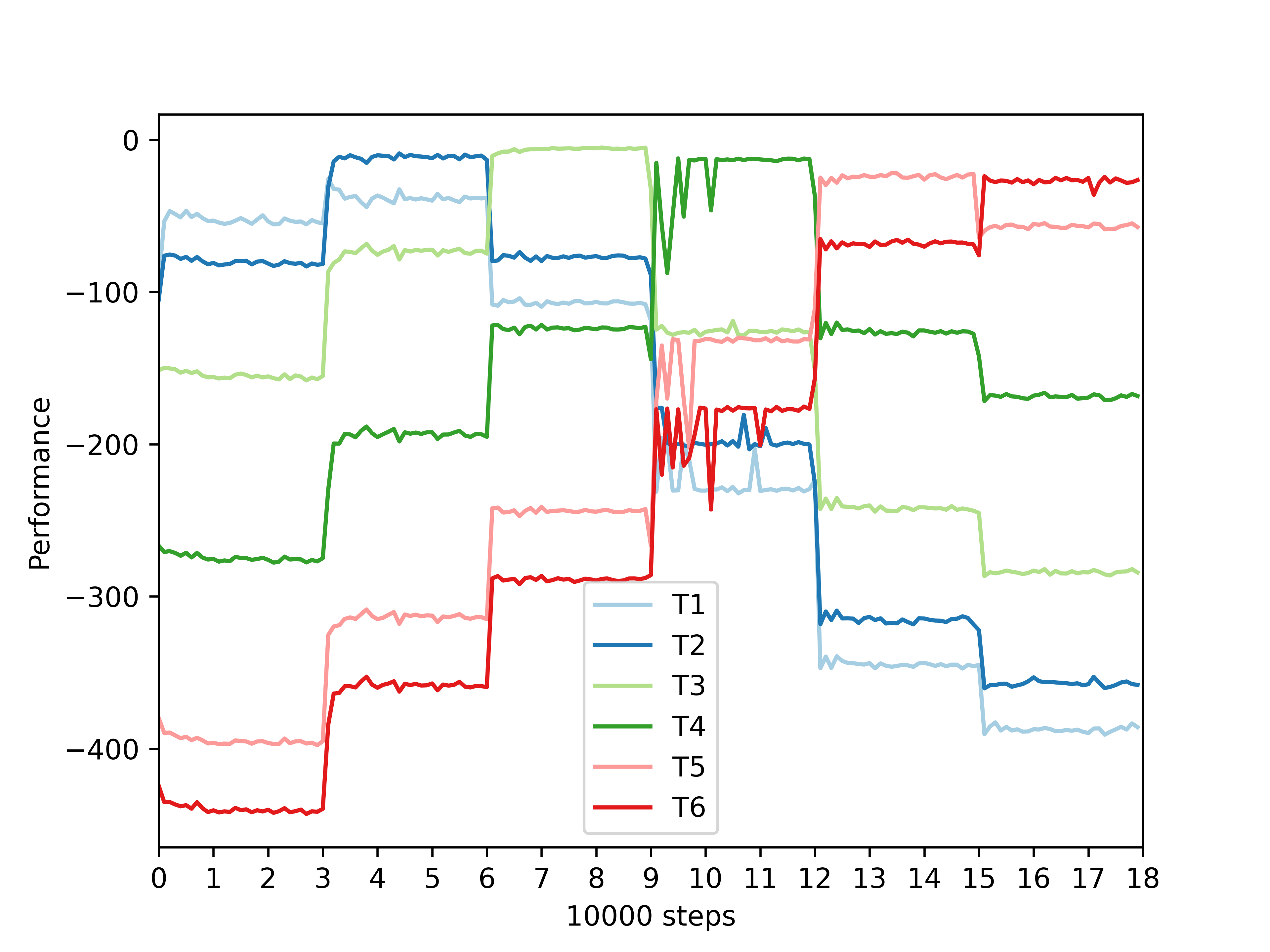}}
  \subcaptionbox{TD3+BC\label{fig:offline}}
    {\includegraphics[width=0.3\linewidth]{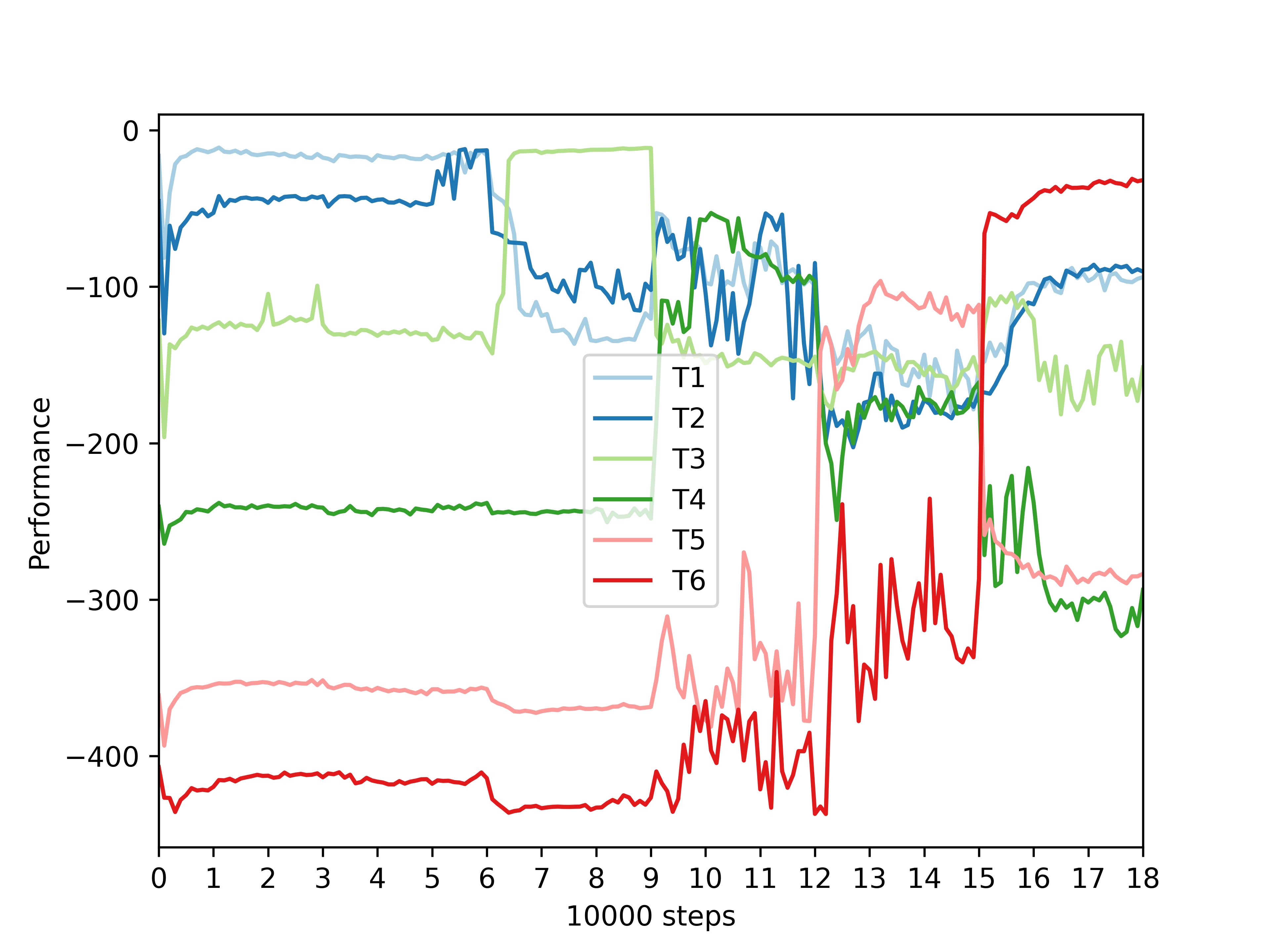}}
  \caption{Performance on each task during continuous learning. The target speed of the cheetah in $T1$ to $T6$ keeps increasing and the task becomes more and more difficult. Each task is most similar to the tasks adjacent to it in sequence.}
  \label{fig:explore}
\end{figure}

\textbf{In CORL setting, what are the advantages of DT compared with AC structure algorithm?} 
% Through the comparison in Fig.\ref{fig:explore}, we find that DT has the several advantages in CORL settings:

\begin{itemize}[leftmargin=*]
    \item DT's learning efficiency is higher than that of AC mode algorithms, and it can learn better-performing policies with the same data quality. This is even more necessary in a continuous learning setting, because in limited training steps, in addition to learning new tasks, operations such as distillation also need to be performed to achieve knowledge transfer. Faster single-task learning efficiency is more conducive to subsequent transfer. 
    \item DT exhibits a distinctive capability in task identification. Notably, when there is a shift in the training dataset, DT demonstrates an immediate adjustment in its performance, corresponding to the change in tasks. This observation underscores its inherent ability for implicit task identification. In contrast, offline algorithms tend to exhibit more erratic learning curves. 
    \item The model's performance can benefit from similar tasks (e.g. Performance on vel-26 keeps improving since the current task is getting closer), which gives us the possibility to review selectively.
    \item DT has a powerful memory ability and generalization ability in disguise, which is exactly what we need in the CORL setting. In steps 90K to 120K in Fig.\ref{fig:explore}, DT has better zero-shot generalization performance in several untrained tasks, which means DT trained on previous offline data can directly perform well on unseen tasks.
\end{itemize}

\textbf{What problems do we need to solve when applying DT to CORL?} DT can identify changes in offline data distribution faster and quickly converge to a better policy under the new task, which also means that the forgetting problem of DT is more serious. We attribute this to the fact that DT is trained through supervised learning, and the vanilla DT updates all network parameters during training. Therefore, DT with history trajectory $\tau$ as input can more accurately perceive changes in the distribution of offline datasets, which are then reflected in changes in action output.
In contrast, AC-structured methods typically follow a two-step process. They initially correct the estimation of Q value using temporal difference learning and subsequently adjust the parameters of the actor network to maximize Q. This process renders AC-based methods less susceptible to changes in the offline data distribution compared to DT.
% In contrast, the method based on the AC structure first needs to correct the critic network estimation based on temporal difference learning, and then change the parameters of the actor network with the objective of maximizing the value function, which causes this method to be less sensitive to changes in offline data distribution than DT.
% At the same time, DT's better zero-shot generalization demonstrates that it contains knowledge that can be shared between tasks. What we need to do is to keep and exploit the knowledge shared between these tasks, while updating and saving task-specific knowledge to implement adaptation and prevent forgetting.
Simultaneously, DT's superior zero-shot generalization capability indicates its capacity for shared knowledge between tasks. Our objective, therefore, is to preserve and harness this shared knowledge across tasks while concurrently updating and preserving task-specific knowledge. This approach facilitates adaptation and guards against forgetting.

% Inspired by the above observation and analysis, we further propose MH-DT and LoRA-DT, two new DT-based methods that leverage the strengths and solve DT’s more serious forgetting problem in CORL problem.

\section{Methodology}
% In this section, we first propose MH-DT in the scenario where replay buffers are available, a method that uses multiple heads to store task-specific knowledge and share knowledge with a common part.  When the replay buffer is unavailable, we propose LoRA-DT, which merges weights that have little impact for knowledge sharing and fine-tunes the decisive MLP layer in DT blocks with LoRA to adapt to the current task. Both methods are based on analysis in Sec.\ref{chp:rethink} and aim to separate shared knowledge and task-specific knowledge. \ls{shouldd be simplified. One or two sentences are enough.}
Inspired by the above observation and analysis in Sec.\ref{chp:rethink}, we further propose MH-DT and LoRA-DT, two new DT-based methods that leverage the strengths and solve DT’s more serious forgetting problem within CORL setting when the replay buffer is available or not respectively.

\subsection{MH-DT: replay buffer based CORL}
Prior rehearsal-based approaches utilize a multi-head policy network $\pi$ and a Q-network $Q_n$ to learn task $T_n$, which means, during learning, policy network $\pi$ has two objectives: on the one hand, the multi-head policy $\pi$ is optimized for all current and previous tasks $T_1$ to $T_{n}$, for action prediction; on the other hand, the policy $\pi$ is also used for updating the current Q-network $Q_n$. This dual role of the policy network can lead to a significant performance decline in the rehearsal phase.
% In order to solve this problem, \citet{Gai_Wang_He_2023} proposed OER to learn a policy $\mu_n$ and the corresponding $Q_n$ when learning the task $T_n$. Subsequently, they perform action cloning for current policy $\pi$ with the actions of $\mu_n$ and the actions stored in the buffer from previous tasks $T_1, \dots, T_{n-1}$ simultaneously.
While \citet{gai2023offline} solves the inconsistency between the learning and the review objectives through the introduction of an intermediate policy $\mu_n$, it also presents a new challenge. Directly cloning multiple action distributions into a multi-head policy is meaningless and unexplainable because there is a correlation between the Q functions of different tasks, but no obvious relationship between the policy networks, which are designed to maximize Q. 
Experiments demonstrate that although it effectively learns a high-performing $\mu_n$, $\pi$ struggles to learn a policy that performs well on $T_n$ as review tasks increase.
% Experiments also prove that although it can learn a well-performed $\mu_n$ when the number of tasks that need to be reviewed increases, $\pi$ cannot learn a policy that performs well on $T_n$. 
Additionally, the Rehearsal process can lead to performance degradation due to a distribution shift between the saved trajectory and the learned policy.
% In addition, OER will also cause performance degradation during the rehearsal process due to the distribution shift between the saved trajectory and the learned policy.

\begin{figure}[htbp]
  \centering
  {\includegraphics[width=0.8\linewidth]{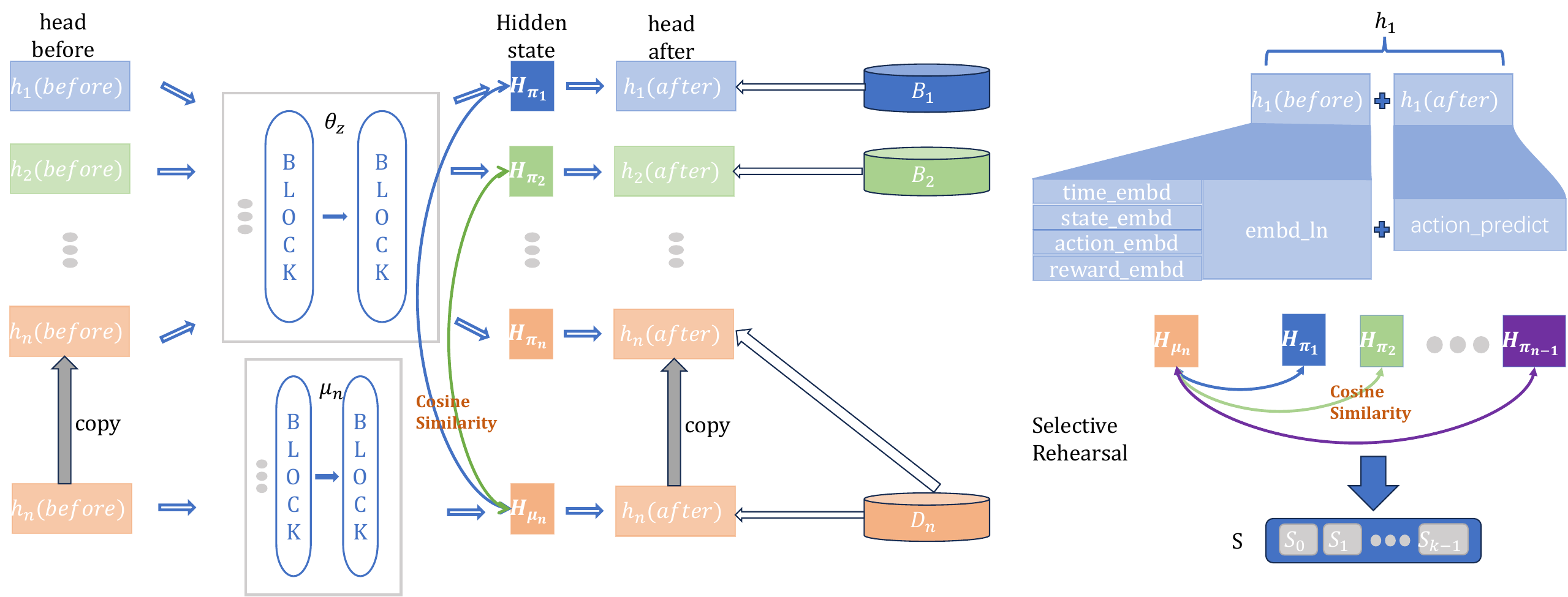}}
  \caption{Schematic diagram of MH-DT. 
  % $\pi_n$ is defined to represent the network with joint parameters $[\theta_z, \theta_n]$ for task $T_n$. 
  The left part is the training process. We first learn a separate policy $\mu_n$, copy the parameters of the head part to head $h_n$, then calculate the loss in Eq.(\ref{eq:loss}) through the data in replay buffer $B_1, \dots, B_{n-1}$ and $D_n$, and update the corresponding head and shared parameters. The upper right part is the structure of each head. The front part includes embedding layers and a layer-norm layer, and the back part includes a linear layer for predicting actions. The lower right part is the schematic diagram of task selection through cosine similarity.}
  \label{fig:mhdt}
\end{figure}

% Based on such analysis, we propose multi-head Decision Transformer (MH-DT). This method uses DT to bypass the process of learning the Q function and avoids the problem of wrong estimation of the Q value in offline settings. And since we directly use supervised learning for training, we do not need to consider the distribution shift between the learned policy and saved data, so we can directly save a part of the previous offline dataset $D_n$ as a replay buffer $B_n$. 
Based on this analysis, we introduce Multi-Head Decision Transformer (MH-DT). This approach leverages DT to circumvent the Q-function learning step, mitigating issues associated with inaccurate Q-value estimation in offline settings.
By employing supervised learning directly for training, we eliminate the need to address the distribution shift between the learned policy and saved data. Consequently, we can readily designate a portion of the prior offline dataset, $D_n$, as a replay buffer, denoted as $B_n$.
The schematic diagram of our proposed architecture is given in Fig.\ref{fig:mhdt}. The intermediate transformer module is used to learn shared environment knowledge, whose parameters are denoted as $\theta_z$. Each task possesses its dedicated head $h_n$ to store task-specific information whose parameters are denoted as $\theta_{n}$. Each head $h_n$ comprises two components: embedding layers and a layer-norm layer before the common module, plus a linear network responsible for action prediction following the common module. We denote $\pi_n$ as the network with combined parameters $[\theta_z, \theta_n]$ specific to task $T_n$ for evaluation. $\pi$ represents the entire MH-DT. 

% Only the policy $\pi_i$ with parameters $[\theta_z, \theta_i]$ are used to evaluate on task $T_i$.

Specifically, when training task $T_n$, we first train a DT policy $\mu_n$ separately and copy the parameters in the head of $\mu_n$ directly to the corresponding head $\theta_n$. Then, the entire MH-DT is updated by the three-part objective. The first is the action prediction goal of DT.
\begin{equation}
\mathcal{L}_{\text {predict }}=\mathbb{E}_{\tau \sim \mathcal{D}_n}\left(\pi_n(\tau)-a_{target}\right)^2
\end{equation}
where $\tau$ is a trajectory sequence from $D_n$ and $a_{target}$ is the last action in $\tau$. Secondly, we use a \textbf{distillation objective} to force $\pi_n$ to be close to $\mu_n$ to enhance learning ability. 
\begin{equation}
\mathcal{L}_{\text {distillation }}=\mathbb{E}_{\tau \sim \mathcal{D}_n}\left(\pi_n(\tau)-\mu_n(\tau)\right)^2 + \mathbb{E}_{\tau \sim \mathcal{D}_n}\left(\displaystyle \mH_{\pi_n}-\displaystyle \mH_{\mu_n}\right)^2
\label{eq:loss_distill}
\end{equation}
where $\displaystyle \mH_{\pi_n}$, $\displaystyle \mH_{\mu_n}$ denote the hidden states of $\pi_n$ and $\mu_n$ networks respectively, which consist of a sequence of hidden vectors. Such additional distillation loss from intermediate states has been shown to improve results in distilling PLMs \citep{jiao2020tinybert}. The last one is a rehearsal objective, aiming to clone the previous experience in $B_1$ to $B_{n-1}$.
\begin{equation}
\mathcal{L}_{\text {rehearsal }}=\frac{1}{n-1} \sum_{j=1}^{n-1} \mathbb{E}_{\tau \sim \mathcal{B}_j}\left(\pi_j(\tau)-a\right)^2
\label{eq:rehearsap}
\end{equation}
% It is worth noting that previous methods based on experience replay have a serious problem, that is, as the number of tasks that need to be reviewed increases, the effect of learning the current task will decrease. To solve this problem, we propose selective review based on the advantage of benefiting from training on similar tasks exhibited by DT as in Sec.\ref{chp:rethink}. 
To address the decreased learning effect caused by multiple review tasks, we propose a \textbf{selective rehearsal} mechanism, capitalizing on DT's ability to benefit from training on similar tasks, as discussed in Section \ref{chp:rethink}.
The task with the lowest similarity is most susceptible to forgetting during the current task's learning process, whereas tasks with greater similarity can benefit from the training on the current task. 
Specifically, instead of reviewing all the previous tasks $T_1$ to $T_{n-1}$, the similarity between the previous tasks and the current task is measured, and only the K tasks with the smallest similarity $[T_{s1},\dots, T_{s_k}]$ are reviewed. 
We collect a batch of data from the current training data set $D_n$, calculate the \textbf{cosine similarity} of the hidden states of each $\pi_j, j = \{1,\dots,n-1\}$ and $\mu_n$. These similarities are then sorted, and the indices of the smallest $k$ tasks are compiled into a list, $S$.
\begin{equation}
S=argsort([C_1, \dots, C_{n-1}])[0:K], \text {where } C_j=\mathbb{E}_{\tau \sim \mathcal{D}_n}Cosine\_Similarity(\mH_{\pi_j}, \mH_{\mu_n})
\label{eq:select}
\end{equation}
$K$ is a hyperparameter that determines the number of tasks that need to be reviewed. Then the rehearsal objective in Eq.(\ref{eq:rehearsap}) can be written as:
\begin{equation}
\mathcal{L}_{\text {rehearsal }}=\frac{1}{k} \sum_{j=s_0}^{s_{k-1}} \mathbb{E}_{\tau \sim \mathcal{B}_j}\left(\pi_j(\tau)-a\right)^2
\label{eq:loss_rehearsal}
\end{equation}
The total loss function for training the whole policy $\pi$ is then,
\begin{equation}
\mathcal{L}_{\text {total }}=\mathcal{L}_{\text {predict }} + \lambda_1 \mathcal{L}_{\text {distillation }} + \lambda_2 \mathcal{L}_{\text {rehearsal }}
\label{eq:loss}
\end{equation}
where $\lambda_1$ and $\lambda_2$ are weights to balance the impact of distilling knowledge about the current task and reviewing previous tasks.

\subsection{LoRA-DT: Replay Buffer Free CORL}
In real-world scenarios, sometimes the replay buffer is not available due to reasons such as data privacy. In order to solve this problem, we propose a method, LoRA-DT, based on the characteristics of DT that does not require a replay buffer to avoid forgetting.

% \citet{Lawson_Qureshi_2023} explored the possibility of directly merging the weights of DTs trained on different tasks to obtain a multi-task model and demonstrated the impact of merging different modules in the DT block (Attention, Layer norm, MLP) on the final performance through experiments. 
\citet{lawson2023merging} investigated the feasibility of directly merging the weights of Decision Transformers (DTs) trained for different tasks to create a multi-task model. 
% They conducted experiments to assess how merging different modules within the DT block, such as Attention, Layer Norm, and MLP, affects the final performance.
Their findings suggest that, for sequential decision-making tasks, DTs rely less on attention and place more emphasis on MLP layers. 
% Experiments indicate that directly averaging the weights of the attention layer from another task with the current task has minimal impact while fusing the weights of MLP results in a significant performance degradation.
% In this paper, they proposed that for sequential decision-making tasks, DTs do not heavily rely on attention but rely more on MLP layers. Experiments also show that directly averaging the weight of the attention layer of another task and the weight of the current task has little impact. On the contrary, fusing the weight of the MLPs will lead to a significant decline in performance. 
Low-Rank Adaptation (LoRA) \citep{hu2021lora}
% is a training method that accelerates the training of large models while consuming less memory. It 
adds pairs of rank-decomposition weight matrices (called update matrices) to existing weights and only trains those newly added weights for efficient tuning and avoiding forgetting.
% Previous pretrained weights are kept frozen so the model is not as prone to catastrophic forgetting. 
\begin{figure}[htbp]
  \centering
  {\includegraphics[width=0.8\linewidth]{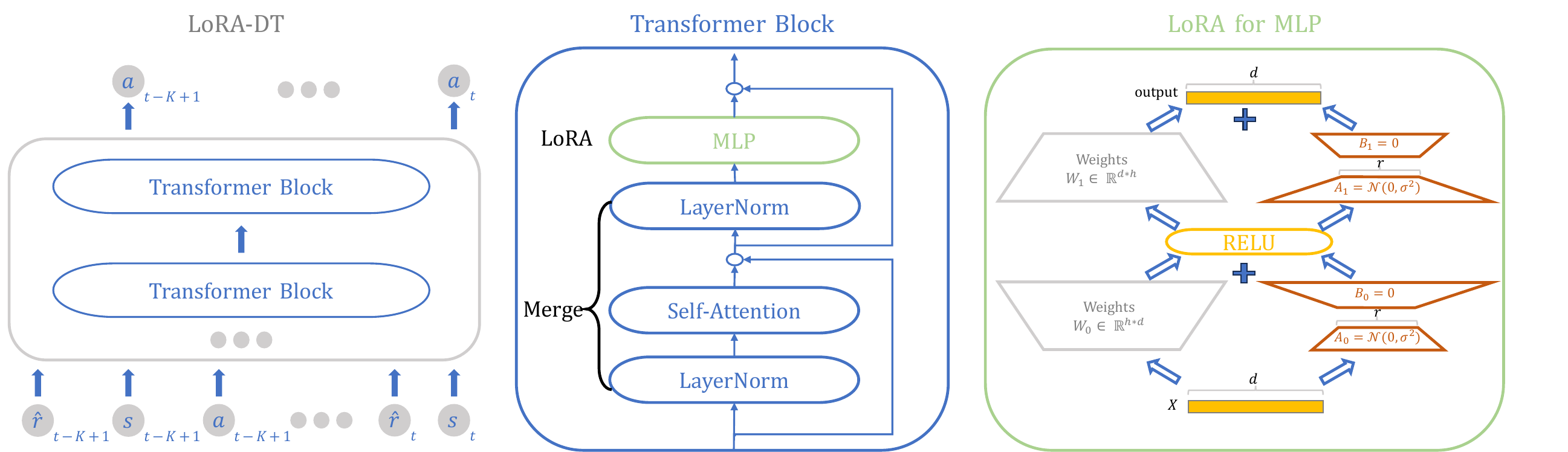}}
  \caption{Model architecture of LoRA-DT. In each block of DT, we first fuse and freeze the weights of layers except the MLP layer as in Eq.(\ref{eq:merge}), then use LoRA to fine-tune the MLP layer as in Eq.(\ref{eq:fine-tune}). The rightmost picture is a schematic diagram of LoRA. We fix the original parameter matrix $\mW_0, \mW_1$, multiply the two matrices $\mA\mB$ to represent the update of the weight matrix, and add it to the original calculation result.}
  \label{fig:lora}
\end{figure}

Inspired by the weight property of DT and saved LoRA matrices can be used to avoid catastrophic forgetting, we proposed LoRA-DT to fine-tune the MLP layer in each block of DT through Low-rank adaptation and save the LoRA matrices $\mA\mB$ of each task. The schematic diagram of our proposed architecture is given in Fig.\ref{fig:lora}.  When rank $r$ is small enough, saving the $\mA\mB$ matrices is significantly more memory efficient than saving the replay buffer, and has better ability to prevent catastrophic forgetting.
% \begin{figure}[!h]
%   \centering
%   \subcaptionbox{Decision Transformer\label{fig:DT}}
%     {\includegraphics[width=0.3\linewidth]{figure/DT.pdf}}
%   \subcaptionbox{TD3+BC\label{fig:offline}}
%     {\includegraphics[width=0.25\linewidth]{figure/block.pdf}}
%     \subcaptionbox{TD3+BC\label{fig:offline}}
%     {\includegraphics[width=0.35\linewidth]{figure/MLP.pdf}}
%   \caption{Performance on each task during continuous learning.}
%   \label{fig:lora}
% \end{figure}
Specifically, when training the first task $T_1$, we train a DT model $\pi$ and update all parameters in it. Then, when training $T_n, n>1$, we first train a DT model $\mu_n$ separately, and fuse all parameters except the MLP layer in each block with the current model $\pi$ through:
\begin{equation}
\theta_{\pi}=(1-\lambda)\theta_{\pi}+\lambda \theta_{\mu_n}
\label{eq:merge}
\end{equation}
where $\theta$ represents all parameters in DT except the MLP parameters of each block and $\lambda$ is a weight to balance the impact of the merge. Then, we fine-tune MLP layers using LoRA.  
% Specifically, the MLP layer in each block in the original DT implementation is composed of two Cov1D layers plus an intermediate activation layer. In our implementation, we replace the Cov1D layer with a liner class \footnote{https://github.com/microsoft/LoRA} which contains a LoRA Layer and a Linear layer. Update the parameters of the Linear layer when training the first task $T_1$, and then only fine-tune the LoRA matrix $\mA\mB$ and save it after training each task. 
The form of the original MLP layer is:
\begin{equation}
MLP(\mX)=\mW_1(RELU(\mW_0\mX+b_0))+b_1
\end{equation}
where $\mW_0 \in \mathbb{R}^{d \times h}$  and $\mW_1 \in \mathbb{R}^{h \times d}$ represent the weight of two linear layers, $b$ is bias and $\mX \in \mathbb{R}^{l \times h}$ is the input of MLP layer. $h, d, l$ respectively represent the hidden dim of the transformer, the inner dim of the MLP layer and the input token length. The form of fine-tuning using LoRA is as follows:
\begin{equation}
MLP(\mX)=(\mW_1+\mB_1\mA_1)RELU((\mW_0+ \mB_0\mA_0)\mX+b_0)+b_1
\label{eq:fine-tune}
\end{equation}
$\mA_0 \in \mathbb{R}^{r \times h}$, $\mB_0 \in \mathbb{R}^{d \times r}$, $\mA_1 \in \mathbb{R}^{r \times d}$ and $\mB \in \mathbb{R}^{h \times r}$ are update matrices in which $r$ is the rank of LoRA and $r<<\min (d, h)$.  
% When training on task $T_n, n>1$, we first initialize $\mA\mB$ to random Gaussian distribution and 0 matrices respectively, then fix $\mW_0, \mW_1$ and start updating the $\mA\mB$ matrices. 
After training, we save the updated matrices $ \sM_n=k * [\mA_0,\mB_0,\mA_1,\mB_1]$ for task $T_n$, where $k$ is the number of blocks in DT. 
The space occupied by saving update matrices for each task is $2* k * r *(h+d)$.
% which is significantly less than the space to save the replay buffer. 
We only need to change the update matrices to the corresponding $\sM_i$ when evaluating task $T_i$.

\section{EXPERIMENTS}
% We conduct extensive experiments to demonstrate the effectiveness of our proposed scheme and test whether we can keep both stability and plasticity when learning sequential offline RL tasks.
\subsection{Setup}
\textbf{Baselines}
% Because \citet{Gai_Wang_He_2023} experimentally proved in their paper that the online continuous reinforcement learning algorithm is not suitable for offline settings and the OER has exceeded all previous algorithms. 
In our experiments, we mainly compare the current SOTA method OER and some DT variants (combination of continual learning methods and DT). Our baselines are as follows:
\begin{itemize}[leftmargin=*]
    \item \textbf{OER} \citep{gai2023offline}: Use a trained model to select experience and a new dual behavior cloning (DBC) architecture to avoid the disturbance of behavior-cloning loss on the Q-learning process of AC structure.  It currently stands as the most proficient method within CORL setting.
    \item \textbf{PDT} (Prompt Decision transformer) \citep{xu2022prompting}: A multi-task method based on DT. 
    % We show the performance on PDT as a reference. 
    The multi-task learning setting can obtain datasets on all tasks simultaneously so that does not suffer from the catastrophic forgetting problem and can be seen as an upper bound on performance.
    \item \textbf{Vanilla DT} \citep{cha2021co2l}: We directly apply vanilla DT to the CORL setting for comparison, in order to prove that our method can indeed reduce catastrophic forgetting.
    \item \textbf{DT + EWC} \citep{kirkpatrick2017overcoming}: a regularization-based approach that calculates the importance of each parameter to the old task and limits the changes of these important parameters when learning a new task, thus maintaining the memory of the old task.
    \item \textbf{DT + SI} \citep{zenke2017continual}: a regularization-based approach that uses the impact of changes in parameters to the change in the loss value to define the importance.
    \item \textbf{DT + GEM} \citep{lopez2017gradient}: a rehearsal-based method using an episodic memory of parameter gradients to limit the DT network update.
    
\end{itemize}

\textbf{Offline Sequential Datasets}
We consider four sets of tasks in two data qualities from environments as in \citet{gai2023offline} and \citet{mitchell2021offline}: Ant-Dir, Walker-Par, Cheetah-Vel and Meta-World reach-v2. For each environment, we randomly sample six tasks to form sequential tasks $T_1$ to $T_6$. To consider different data quality, we selected different time periods in the online training buffer and obtained expert-quality data and middle-quality data as in \citet{mitchell2021offline}. For Meta-World reach-v2, we use the expert dataset because only expert script policies are available.
% \begin{itemize}[leftmargin=*]
%     \item Ant-2D Direction (Ant-Dir): train a simulated ant with 8 articulated joints to run in a 2D direction
%     \item Walker-2D Params (Walker-Par): train a simulated agent to move forward, where different tasks have different parameters. Specifically, different tasks require the agent to move at different speeds
%     \item Half-Cheetah Velocity (Cheetah-Vel): train a cheetah to run at a random velocity. Cheetah-vel is unique in that as the 'vel' number increases the task becomes more challenging.
%     % Cheetah-vel is special because the larger the vel number, the greater the target speed and the more difficult the task.
%     \item Meta-World reach-v2. Tasks are to control a Sawyer robot's end-effector to reach different target positions in 3D space. The agent directly controls the XYZ location of the end-effector.
% \end{itemize}

% For Ant-Dir, Walker-Par and Meta-World reach-v2, we randomly sample six tasks to form sequential tasks $T_1$ to $T_6$. For Cheetah-Vel, we fixedly select the six tasks of vel={3, 6, 9, 12, 15, 18} and train them in order. The difficulty of the six tasks increases in sequence.

% To consider different data quality, we selected different time periods in the online training buffer and obtained expert-quality data and middle-quality data as in \citet{Mitchell_Rafailov_Peng_Levine_Finn_2020}. For Meta-World reach-v2, we use the expert dataset because only expert script policies are available.

\textbf{Metrics} 
% We use the metrics mentioned in Sec.\ref{chp:pre} to measure the average performance, forgetting level and generalization ability of the continuous learning algorithm.
Following \citet{lopez2017gradient}, we adopt the average performance (PER), the backward transfer (BWT) and forward transfer (FWT) as evaluation metrics,
\begin{equation}
    \mathrm{PER}=\frac{1}{N} \sum_{n=1}^N a_{N, n}, \mathrm{BWT}=\frac{1}{N-1} \sum_{n=1}^{N-1} a_{n, n}-a_{N, n}, \mathrm{FWT}=\frac{1}{N-1} \sum_{n=2}^N a_{n-1, n}-\bar{b}_n
\label{eq:metrics}
\end{equation}
where $a_{i, j}$ means the final cumulative rewards of task $j$ after learning task $i$ and $\bar{b}_n$ means the test performance for each task at random initialization. PER Indicates the average performance of the tasks and higher is better; BWT measures the forgetting level and lower is better; FWT demonstrates the generalization ability and higher is better. Lower BWT and higher FWT are preferred when similar PER.

See more details for datasets and metrics in Appendix.\ref{sec:detail}.

\textbf{Implementation details} 
After preparing the offline datasets, we train a DT model in the order of $T_1$ to $T_6$ for each environment. It is worth noting that each DT model is trained from scratch, so the processing of the first task $T1$ is slightly different. For details, please refer to the pseudocode in Appendix.\ref{sec:pseudo-code}. The number of training steps for each task is a hyperparameter. We choose 30K in the implementation like \cite{gai2023offline}, which is completely enough for an AC-structure method (about 5K steps) or DT (1K steps) to achieve optimality on a single task. For each task, we train 30K steps and switch to the next. The result is calculated via five repeated simulations with different numbers of seeds. For each evaluation step, we test all models on the corresponding tasks 10 times and report the average.

\subsection{Overall Results}
In this section, we list the performance metrics of all methods under four environments and two data qualities in Table~\ref{tab:metrics}. In addition, we also drew the learning curve of each method to more intuitively observe the learning efficiency and forgetting degree in Fig.~\ref{fig:lr}. Due to space limitations, we show the training curve of Ant\_Dir in the text. See Appendix.\ref{sec:exp} for all experiment results.

\begin{table}[htbp]
    \centering
    \renewcommand\arraystretch{1.2}
    \large
    \caption{Performance of MH-DT and LoRA-DT compared with baselines. For each metric, we highlight the two best-performing methods except the upper bound PDT. PDT has access to all data simultaneously, so its FWT and BWT are not evaluated. }
    \resizebox{\textwidth}{!}{
    \begin{tabular}{c|c|c|c|c|c|c|c|c|c|c|c|c|c}
\hline \multirow{2}{*}{ Dataset Quality } & \multirow{2}{*}{ Methods } & \multicolumn{3}{c|}{ Ant-Dir } & \multicolumn{3}{c|}{ Walker-Par } & \multicolumn{3}{c|}{ Cheetah-Vel } & \multicolumn{3}{c}{ Meta-World reach-v2 }\\
\cline { 3 - 14 } & & PER$\uparrow$ & BWT$\downarrow$ &FWT$\uparrow$& PER$\uparrow$ & BWT$\downarrow$ &FWT$\uparrow$& PER$\uparrow$ & BWT$\downarrow$&FWT$\uparrow$ & PER$\uparrow$& BWT$\downarrow$&FWT$\uparrow$\\
\hline 
\multirow{8}{*}{ Middle } & PDT & 347.9 & - & - & 540.2 & - & - & -39.3 & - & - & - & - & -\\
& OER & 193.6 & 125.6 & -111.8 & 56.0 & 132.9 & 97.3 & -113.8 & 43.8 & 85.1 & - & - & - \\
\cline{ 2 - 14 }
& Vanilla DT & 146.7 & 248.7 & 102.2 & 380.9 & 166.1 & \textbf{331.9} & -118.8&99.5 &\textbf{154.1} & -& -& -\\
& DT + EWC & 140.0 & 245.4 & 89.7 & 367.4 & 158.7 & 238.8 & -120.6&105.1 &169.7 & -& -& -\\
& DT + SI & 152.3 & 207.8 & 109.2 & 326.5 & 196.3 & 304.6 & -141.3&58.2 &130.3 & -& -& -\\
& DT + GEM & 152.5 & 154.4 & \textbf{177.5} & \textbf{445.3} & 76.8 & 295.7 & -113.4&80.7 &\textbf{153.8} & -& -& -\\

\cline{ 2 - 14 }
& MH-DT(ours) & \textbf{326.5} & \textbf{38.3} & \textbf{110.5} & \textbf{510.8} & \textbf{-32.0} & 314.2 & \textbf{-35.4} & \textbf{3.4} & 144.6 & - & - & - \\
& LoRA-DT(ours) & \textbf{230.3} & \textbf{5.1} & 98.4 & 424.6 & \textbf{-2.5} & \textbf{321.1} & \textbf{-57.8} & \textbf{3.6} & 150.4 & - & - & - \\
\hline \multirow{8}{*}{ Expert } & PDT & 528.8 & - & - & 537.2 & - & - & -17.5 & - & - & 532.3 & - & -\\
& OER & 126.0 & 260.7 & -126.5 & 74.0 & 117.5 & 28.3 & -201.9 & 35.1 & -32.4 & 0.6 & 0.4 & 0.4 \\
\cline{ 2 - 14 }
& Vanilla DT & 141.7 & 459.7 & 153.8 & 387.4 & 252.2 & \textbf{333.1} &-119.8 &130.9 &183.8 &90.0 &541.1 &\textbf{2.3}\\
& DT + EWC & 142.0 & 466.8 & 133.6 & 384.5 & 176.3 & \textbf{284.4} &-120.1 &117.8 &168.2 &93.3 &527.7 &1.5 \\
& DT + SI & 154.0 & 374.4 & 113.2 & 244.1 & 166.9 & 242.4 &-137.2 &15.6 &75.9 &104.4 &513.4 &0.5\\
& DT + GEM & 75.1 & 159.1 & 36.7 & 357.3 & 74.7 & 162.3 &-95.5 &91.3 &159.1 &\textbf{428.4} &\textbf{154.9} &1.5\\

\cline{ 2 - 14 }
& MH-DT(ours) & \textbf{437.8} & \textbf{90.8} & \textbf{155.2} & \textbf{505.4} & \textbf{50.4} & 250.8 & \textbf{-21.9} & \textbf{12.9} & \textbf{206.3} & \textbf{441.6} & \textbf{124.5} & 1.0\\
& LoRA-DT(ours) & \textbf{355.1} & \textbf{6.4} & \textbf{150.9} & \textbf{418.6} & \textbf{7.1} & 273.2 & \textbf{-37.3} & \textbf{11.5} & \textbf{211.3} & 150.2 & 248.9 & \textbf{2.0}\\
\hline
\end{tabular}
}
    \label{tab:metrics}
\end{table}

Consistently across tasks and data quality, MH-DT outperforms other algorithms for most test experiments in metrics of PER and FWT while LoRA-DT performs outstandingly on the BWT, showing its strong ability to avoid forgetting. Specifically, we can draw the following conclusions.

For the DT structure, all DT-based methods have higher FWT than AC-based method OER, indicating that the DT structure has stronger zero-shot generalization ability when dealing with similar tasks. In some environments such as Ant-Dir, Vanilla DT has a similar PER with OER, while having significantly larger BWT, indicating a better learning effect and more serious catastrophic forgetting. Directly applying the continuous learning method to DT can alleviate the catastrophic forgetting problem of DT to a small extent. The improvement of the rehearsal-based method (DT + GEM) is relatively larger than regularization-based methods (DT + EWC and DT + SI) which proves that the rehearsal-based continuous learning algorithm is more suitable for CORL settings.

As in the comparison in the first row of Fig.~\ref{fig:lr}, when the regularization-based continual learning method is directly used on DT, there is basically no way to avoid forgetting (the learning curve is the same as Vanilla DT), while the rehearsal-based method reduces the forgetting problem to some extent. Our method MH-DT and LoRA-DT significantly reduces the degree of forgetting, as shown in Fig.\ref{fig:lr_mhdt} and Fig.\ref{fig:lr_lora}.
For MH-DT, while keeping the FWT value almost the same as Vanilla DT, MH-DT significantly reduces BWT and thus improves the PER metric, indicating that MH-DT retains the generalization ability of DT and solves catastrophic forgetting by dividing all models into common parts and task-specific parts. In the most difficult Meta-World environment, MH-DT can also learn better-performing strategies, while OER cannot learn any usable strategies at all. It is also exciting to find that MH-DT can achieve comparable PER  to the upper bound PDT.
\begin{figure}[htbp]
	\centering
	\subcaptionbox{Vanilla DT\label{fig:lr_vandt}}
    {\includegraphics[width=0.245\linewidth]{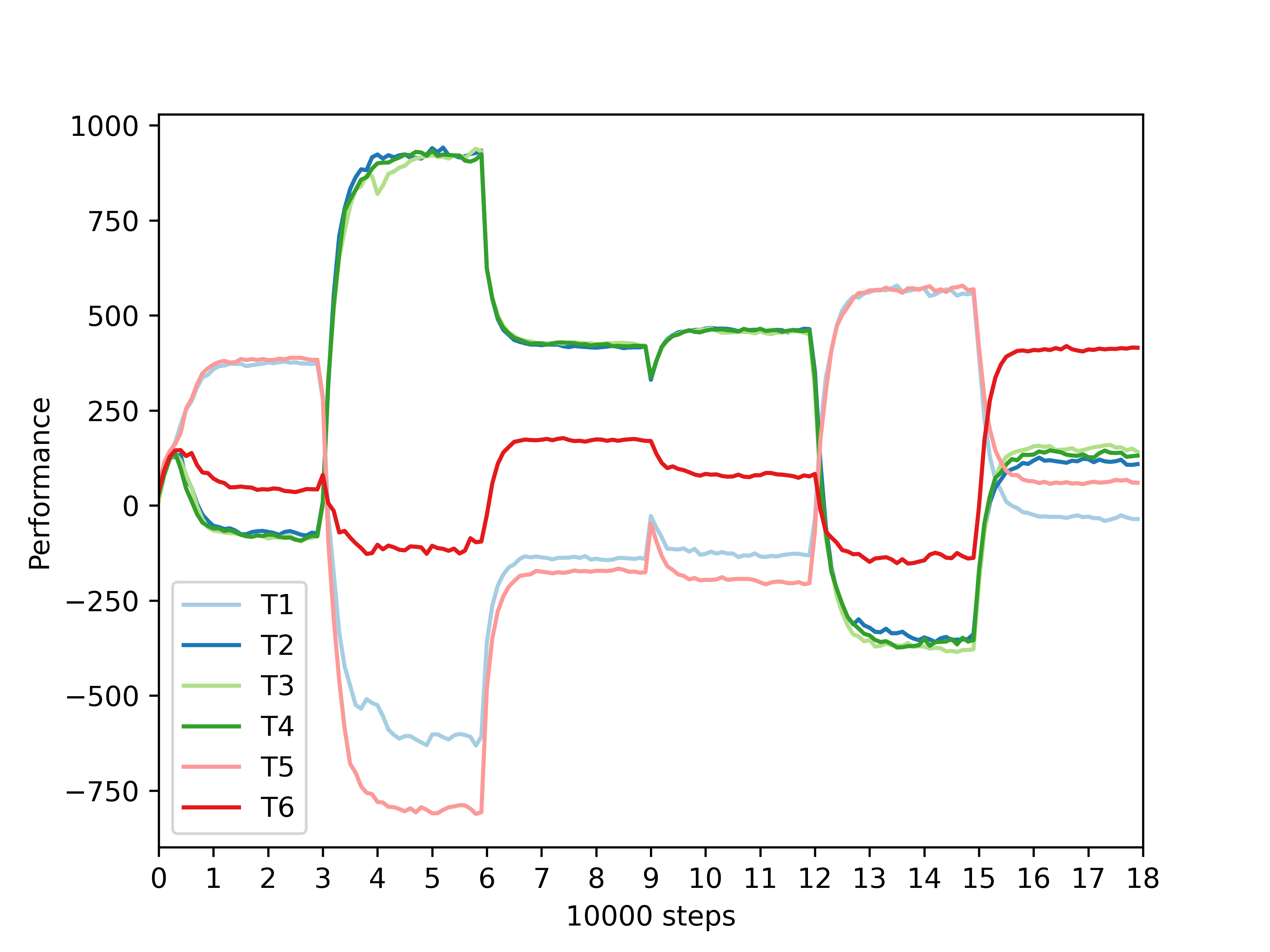}}
    \subcaptionbox{DT + EWC\label{fig:lr_ewc}}
    {\includegraphics[width=0.245\linewidth]{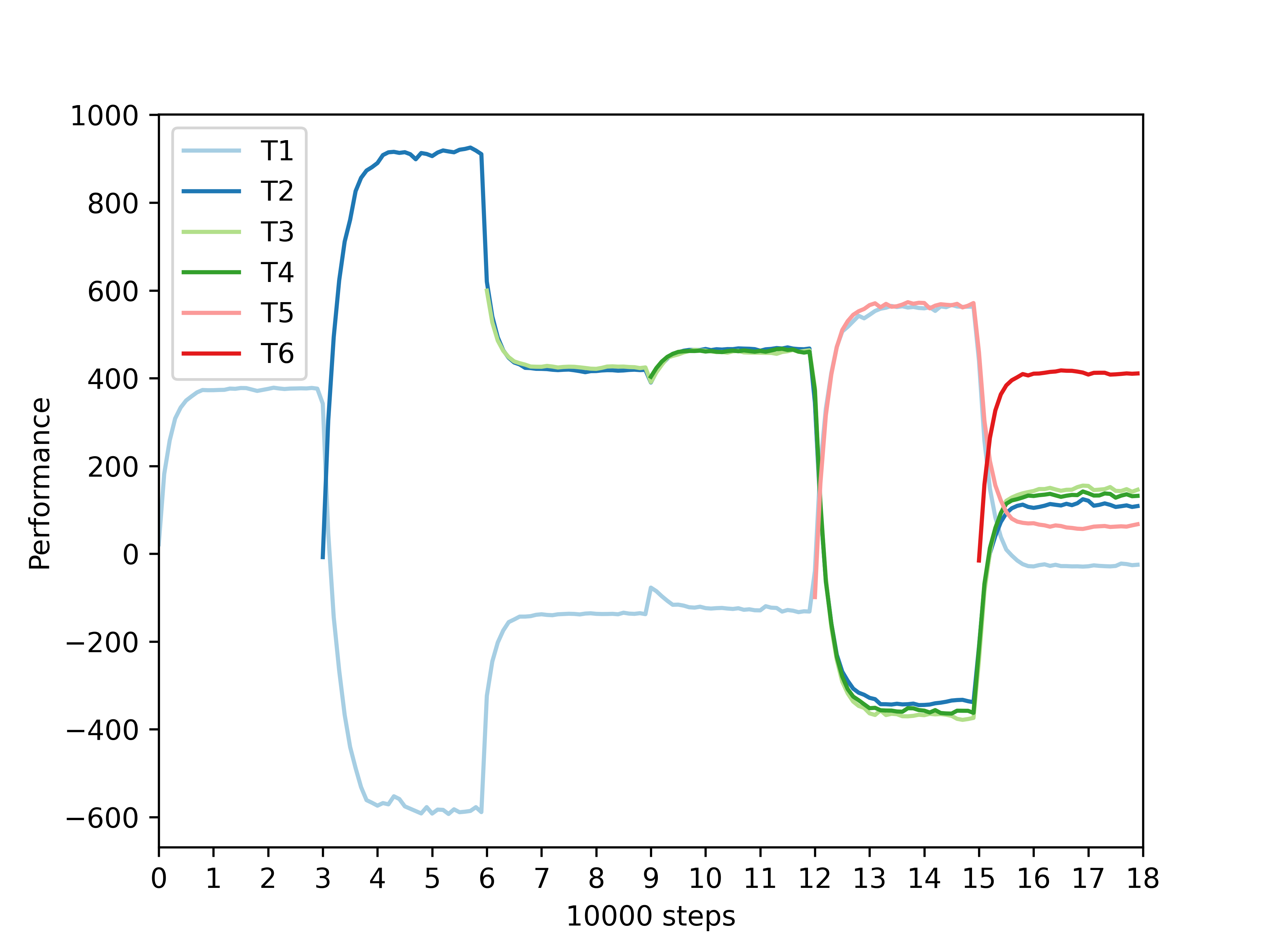}}
    \subcaptionbox{DT + SI\label{fig:lr_si}}
    {\includegraphics[width=0.245\linewidth]{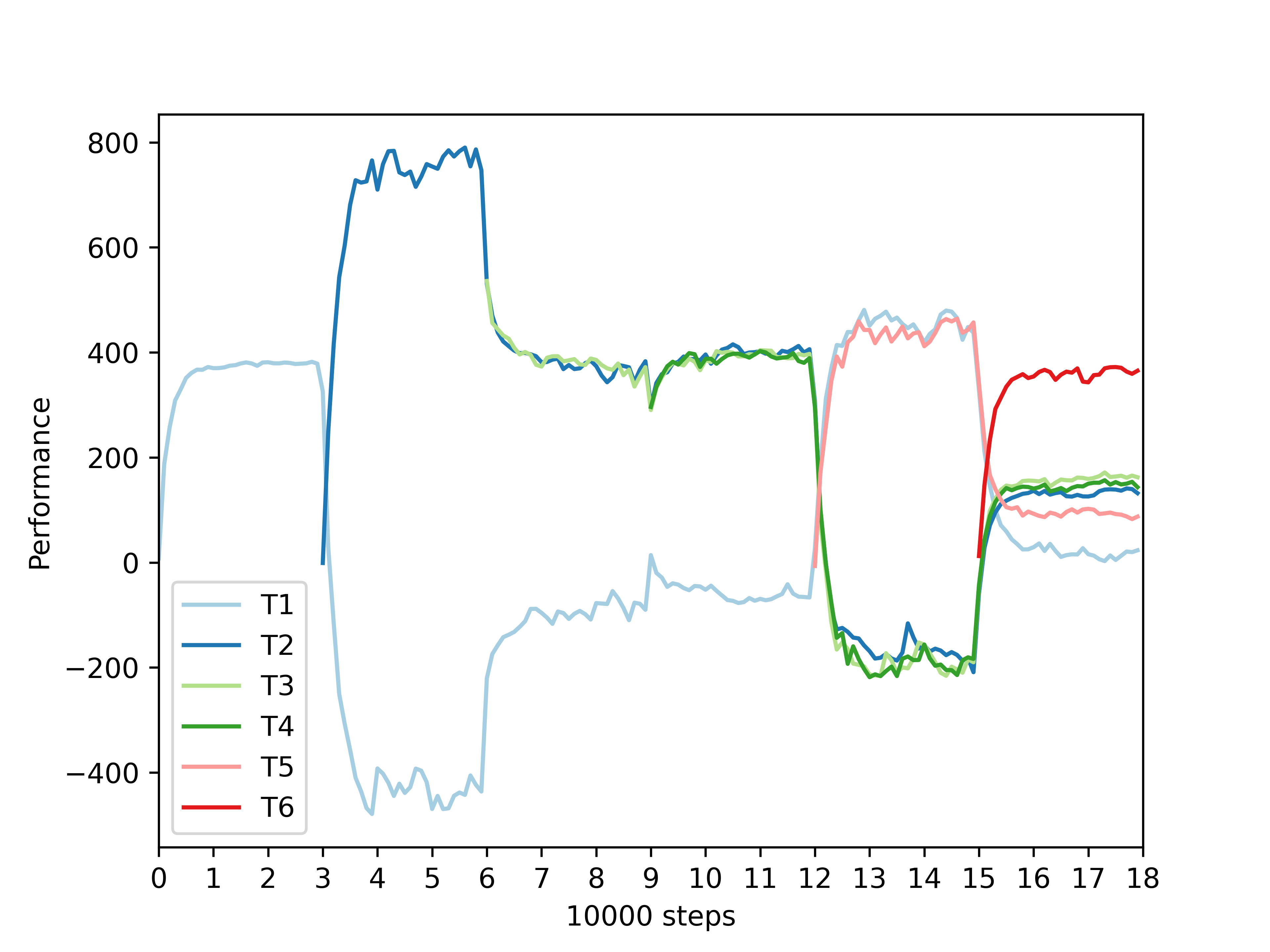}}
    \subcaptionbox{DT + GEM\label{fig:lr_gem}}
    {\includegraphics[width=0.245\linewidth]{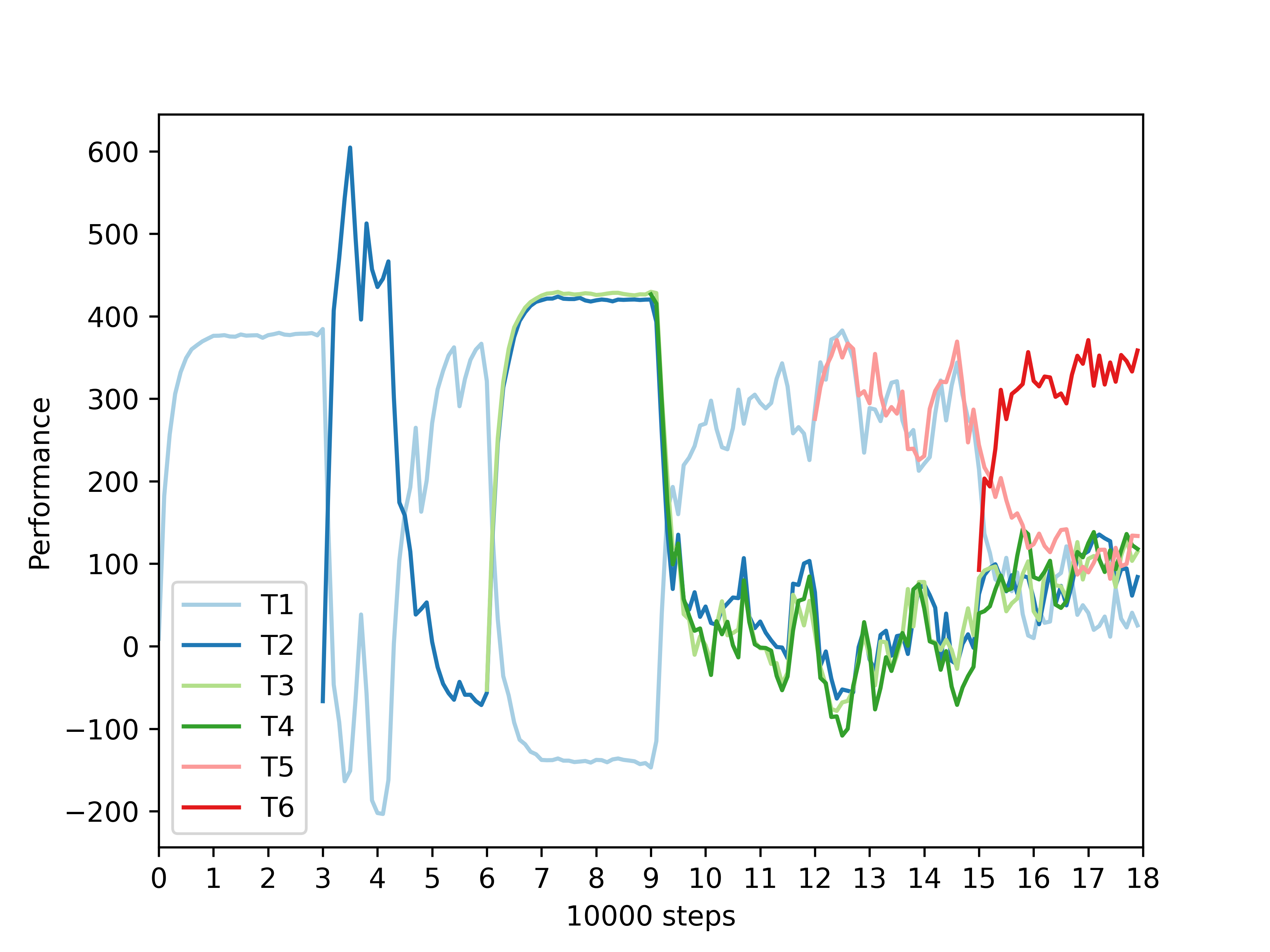}}
	
     \centering	
     \subcaptionbox{PDT\label{fig:lr_pdt}}
    {\includegraphics[width=0.245\linewidth]{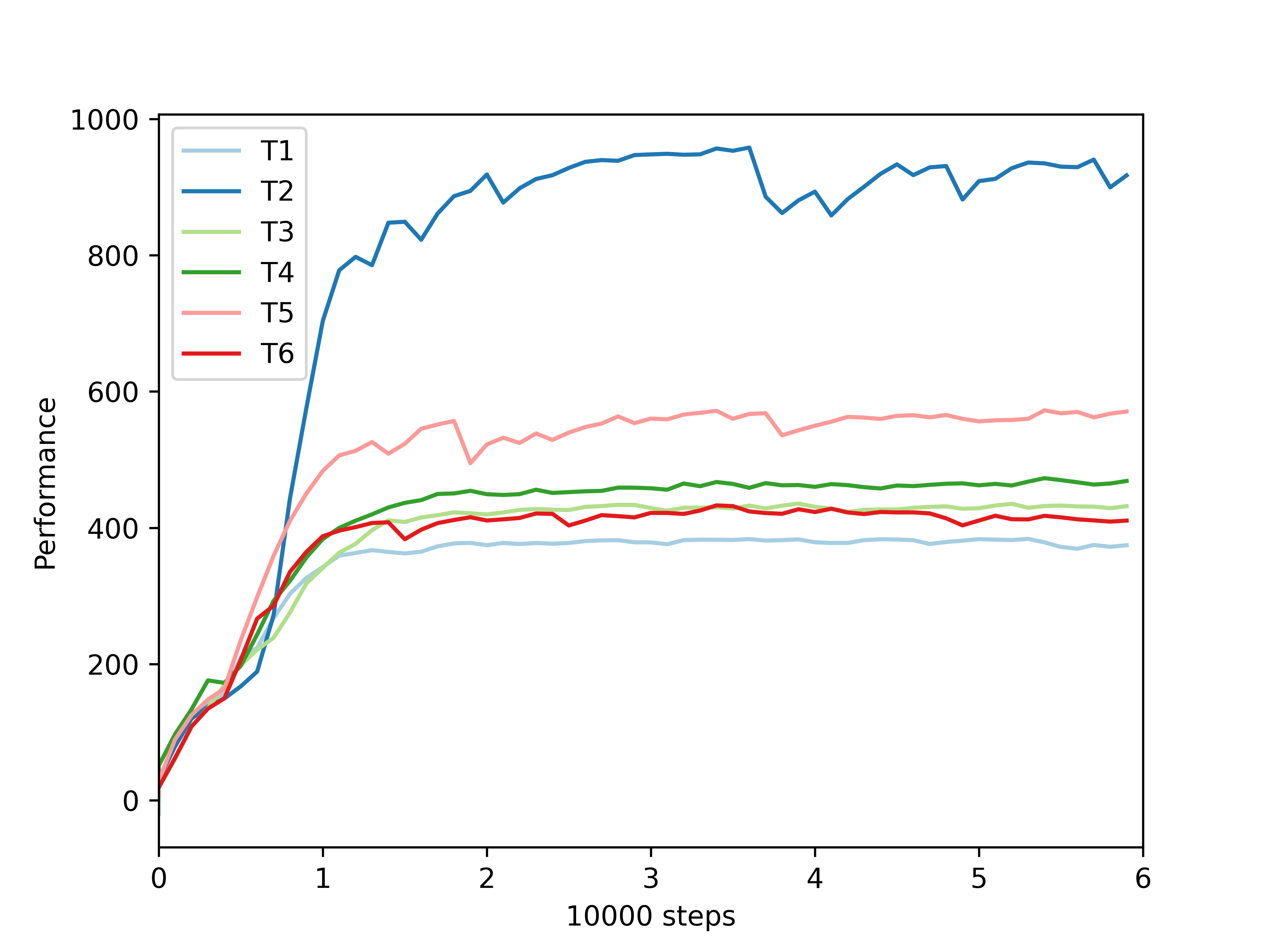}}
        \subcaptionbox{OER\label{fig:lr_oer}}
    {\includegraphics[width=0.245\linewidth]{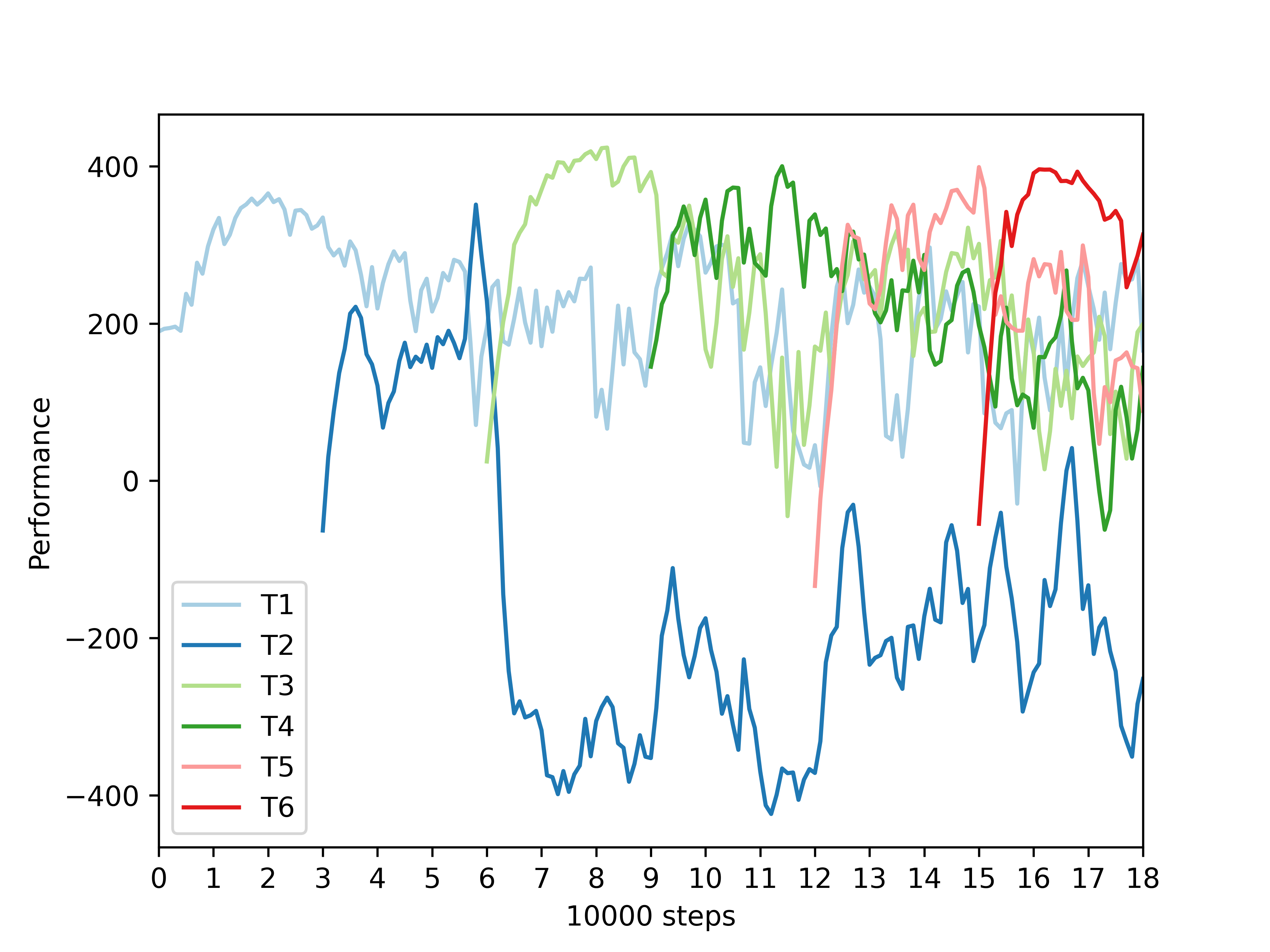}}
    \subcaptionbox{MH-DT\label{fig:lr_mhdt}}
    {\includegraphics[width=0.245\linewidth]{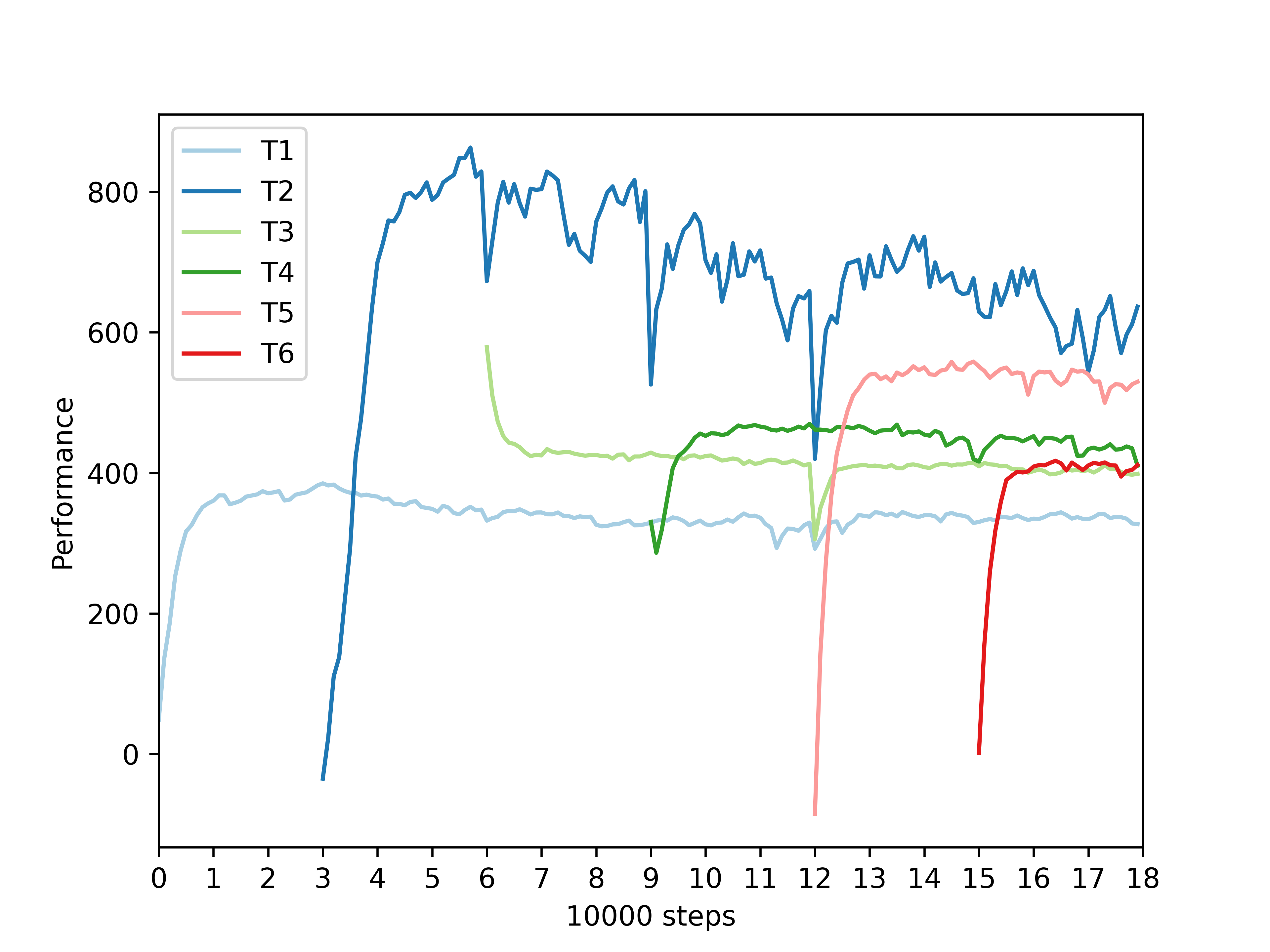}}
	\subcaptionbox{LoRA-DT\label{fig:lr_lora}}
    {\includegraphics[width=0.245\linewidth]{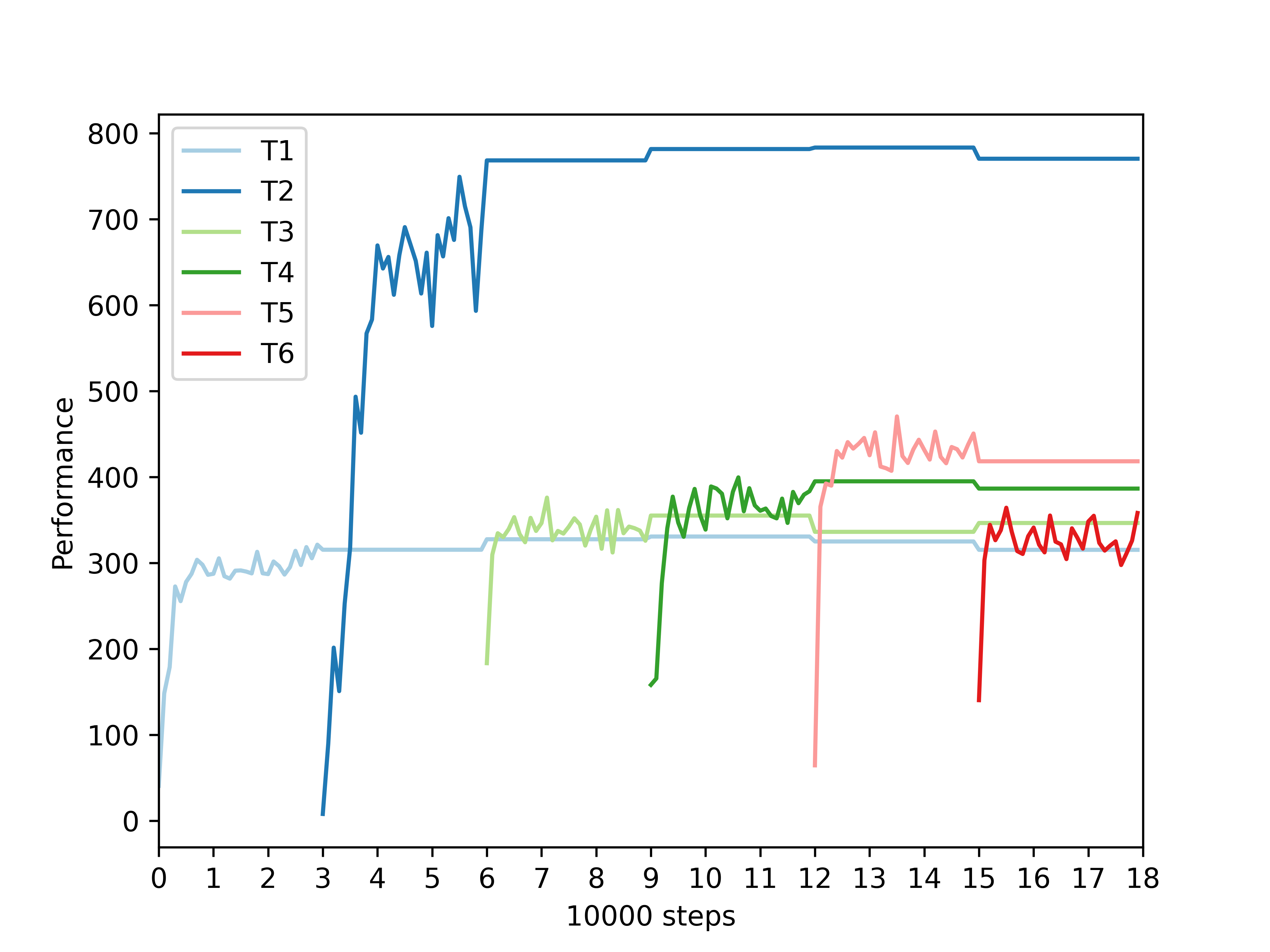}}
    \caption{Process of learning six sequential tasks in Ant\_Dir, where our methods MH-DT and LoRA-DT are compared with five baselines and an upper bound PDT. We train 30K steps on one task for continual learning methods.}
    \label{fig:lr}
\end{figure}

For LoRA-DT, as in Fig.\ref{fig:lr_lora}, the performance of LoRA-DT in previous tasks after the training phase is completely unchanged except for a small fluctuation in the merge phase at the beginning of each new task. 
LoRA-DT demonstrates the highest BWT level in all environments, indicating its exceptional ability to prevent forgetting. LoRA-DT also has better FWT performance, thanks to it sharing most of the network weights between multiple tasks.

% However, even though LoRA-DT demonstrates the highest BWT level, indicating its exceptional ability to prevent forgetting, its performance in PER does not surpass that of MH-DT. This can be attributed to the fact that the fine-tuning method, merge+LoRA, is still not on par with the approach of updating all parameters.

When comparing MH-DT and LoRA-DT, we should first clarify that the setting of LoRA-DT is more strict in which the replay buffer is not available. LoRA-DT's performance in PER does not surpass that of MH-DT, and this can be attributed to the fact that the fine-tuning method, merge + LoRA, is still not on par with the approach of updating all parameters. MH-DT has poorer BWT because each task has its own head and the intermediate processing GPT module is universal. There are errors in encoding and decoding and mismatches as the universal module is updated. LoRA-DT uses an adapter to adapt inaccurate predictions to new tasks and then saves this adaptation module, which is more accurate and less forgotten. The difference in BWT and PER performance between MH-DT and LoRA-DT also reflects the trade-off between plasticity and stability. MH-DT focuses more on the ability to learn new tasks (plasticity), while LoRA-DT focuses more on stability due to its stricter settings. It's also worth noting that in an environment where the FWT is larger, signifying greater task similarity,  the performance of LoRA-DT is not much behind that of MH-DT as in Walker\_Par and Cheetah\_Vel, but relatively larger in other environments.

\subsection{Ablation Study}
\subsubsection{Distillation objective and Selectively Rehearsal in MH-DT}
In MH-DT, we propose an additional distillation objective and selective rehearsal to make the current policy $\pi_n$ closer to our learned teacher policy $\mu_n$, which also means getting better performance on the current task. 
In order to verify the effectiveness of the two modules, distillation objective (DO) and selective rehearsal (SR), we do ablation studies by comparing MH-DT with OER and three variants: MH-DT without distillation objective (MH-DT w/o DO), MH-DT without selective rehearsal (MH-DT w/o DO), MH-DT without distillation objective and selective rehearsal (MH-DT w/o DO w/o SR) on distillation gap (DG), a new metric to measure the gap between the task-specific policy $\mu_n$ that is trained separately and the current policy $\pi_n$ after training process on task $T_n$.
% In order to measure the ability to learn the current task, we propose a new metric, the distillation gap (DG), to measure the gap between the task-specific policy $\mu_n$ that is trained separately and the current policy $\pi_n$ after training process on task $T_n$. It can be expressed as:
\begin{equation}
    \mathrm{DG}=\frac{1}{N} \sum_{n=1}^N t_n-a_{n, n}
    \label{eq:dg}
\end{equation}
where $t_n$ is the final cumulative reward of $\mu_n$, and $a_{n,n}$ is the performance of task $n$ after learning task $n$. A smaller DG indicates a smaller gap and a stronger ability to learn the current task. 
\begin{table}[htbp]
    \centering
    \renewcommand\arraystretch{1.0}
    \tiny
    \caption{Ablation on distillation objective (DO) and selective rehearsal (SR) in MH-DT}
    \resizebox{0.8\textwidth}{!}{
    \begin{tabular}{c|c|c|c|c|c|c}
\hline  \multirow{2}{*}{ Methods } & \multicolumn{2}{c|}{ Ant-Dir } & \multicolumn{2}{c|}{ Walker-Par } & \multicolumn{2}{c}{ Cheetah-Vel } \\
\cline { 2-7 } & PER$\uparrow$ & DG$\downarrow$ & PER$\uparrow$ & DG$\downarrow$ & PER$\uparrow$ & DG$\downarrow$\\
\hline  OER   & 193.6 & 53.3 & 56.0 & 417.9 & -113.8 & 26.5 \\
\hline  MMH-DT w/o DO w/o SR  & 223.9 & 41.1 & 374.8 & 231.7 & -86.1 & 9.8 \\
MH-DT w/o DO   & 310.3 & 30.3  & 456.2 & 90.5 & -54.7 & -4.1 \\
  MH-DT w/o SR   & 302.5 & 37.3 & 423.0 & 125.2 & -64.1 & 0.3 \\
\hline  MH-DT   & 326.5 & -4.0 & 510.8 & 54.1 & -35.4 & -21.6 \\
\hline
\end{tabular}
}
    \label{tab:ab_mhdt}
\end{table}

From Table~\ref{tab:ab_mhdt}, we can observe that: 1)DO and SR can both improve MH-DT's ability to learn current tasks. 2) The DT-based method works better than the AC-based method When using action cloning to transfer knowledge. 3)By correctly selecting reviewed tasks, the performance of the current task can even exceed that of the teacher policy $\mu_n$ as in Cheetah\_Vel.

Thanks to the DT structure, our MH-DT model can utilize these two methods to enhance its learning capabilities. The distillation objective and the similarity used in selective rehearsal are both calculated in the hidden state space of the DT.

\subsubsection{Rank r and Buffer Size}
We compare the space occupied performance and metrics performance of OER, MH-DT and LoRA-DT under different rank $r$ and different sizes of replay buffer. 
% In MH-DT, the size of buffer $B_n$ is selected as 1 K for all $T_n$ similar to \citep{Gai_Wang_He_2023}. And the rank $r=4$ in LoRA-DT.  
In this section, we change the buffer size to 1K, 3K, and 10K and rank $r$ to 4, 16, 64. In order to intuitively compare the size of the occupied space, we take the space of 10K sample buffer as the benchmark and record it as 100\%. It is worth noting that because different environments have different state-action dimensions, the proportion of memory occupied by LoRA-DT is also different.
\begin{table}[htbp]
    \centering
    \caption{Performance when buffer size and rank change}
    \renewcommand\arraystretch{1.0}
    \large
    \resizebox{\textwidth}{!}{
    \begin{tabular}{c|c|c|c|c|c|c|c|c|c|c|c|c}
\hline  \multirow{2}{*}{ Methods } & \multicolumn{3}{c|}{ Ant-Dir } & \multicolumn{3}{c|}{ Walker-Par } & \multicolumn{3}{c|}{ Cheetah-Vel } & \multicolumn{3}{c}{ Meta-World reach-v2 }\\
\cline { 2 - 13 } & PER$\uparrow$ & BWT$\downarrow$ &Memory$\downarrow$& PER$\uparrow$ & BWT$\downarrow$ &Memory$\downarrow$& PER$\uparrow$ & BWT$\downarrow$ &Memory$\downarrow$ & PER $\uparrow$ & BWT$\downarrow$ &Memory $\downarrow$\\
\hline  OER(1K)   & 126.0 & 260.7 & 10\% & 74.0 & 117.5 & 10\% & -201.9 & 35.1 & 10\% & 0.6 & 0.4 & 10\%\\
\hline  OER(3K)   & 150.2 & 234.2 & 30\% & 80.5 & 114.3 & 30\% & -184.2 & 31.1 & 30\% & 0.6 & 0.4 & 30\%\\
\hline  OER(10K)   & 187.3 & 198.5 & 100\% & 96.2 & 110.5 & 100\% & -169.1 & 24.2 & 100\% & 0.6 & 0.4 & 100\%\\
\hline  MH-DT(1K)   & 437.8 & 90.8 & 10\% & 505.4 & 50.4 & 10\% & -21.9 & 12.9 & 10\% & 441.6 & 124.5 & 10\%\\
\hline  MH-DT(3K)   & 454.1 & 74.3 & 30\% & 515.7 & 32.1 & 30\% & -18.4 & 6.3 & 30\% & 463.8 & 100.2 & 30\%\\
\hline  MH-DT(10K)   & 490.6 & 43.2 & 100\% & 537.2 & 12.5 & 100\% & -16.3 & 2.2 & 100\% & 500.3 & 88.1 & 100\%\\
\hline  LoRA-DT(r=4)   & 355.1 & 6.4 & 1.6\% & 418.6 & 7.1 & 2.4\% & -37.3 & 11.5 & 2.2\% & 150.2 & 148.9 & 1.3\%\\
\hline  LoRA-DT(r=16)   & 387.4 & 6.7 & 6.6\% & 427.3 & 6.9 & 9.8\% & -30.5 & 10.3 & 8.7\% & 162.8 & 132.2 & 5.4\%\\
\hline  LoRA-DT(r=64)   & 407.54 & 5.8 & 26.5\% & 469.2 & 7.1 & 39.3\% & -33.4 & 11.3 & 35.1\% & 168.5 & 149.5 & 21.8\%\\
\hline
\end{tabular}
}
    \label{tab:ab_lora}
\end{table}

The results in Table~\ref{tab:ab_lora} demonstrate the memory efficiency of LoRA-DT, it can achieve performance that exceeds OER and is close to MH-DT while using nearly one-tenth of the space used by them. In addition, for methods that are based on ER, a larger buffer size can significantly reduce the BWT value, thereby improving performance. When the size of the replay buffer approaches the training dataset, it will become a multi-task problem. On the contrary, LoRA-DT's forgetting metric BWT is not sensitive to the rank $r$. However increasing $r$ can increase the plasticity of the LoRA-DT model, allowing it to obtain better-performing policies through fine-tuning. When $r$ increases to the inner dim of the MLP, the tuning effect is equivalent to directly fine-tuning the MLP layer.
\section{CONCLUSION}
In this work, we propose DT can serve as a more suitable offline
continuous learner by rethinking the CORL problem, in which we highlight the advantages of DT over AC-structured algorithms and then focus on addressing the more severe forgetting issue. Subsequently, we introduce MH-DT which employs multiple heads to store task-specific knowledge, facilitates knowledge sharing with a common component, and incorporates distillation and selective rehearsal modules to enhance learning capacity.
When replay buffers are unavailable, we propose LoRA-DT, which merges impactful weights for knowledge sharing and fine-tunes the crucial MLP layer within DT blocks using LoRA.  Experiments and analysis show that our DT-based methods outperform SOTA baselines on various continuous control tasks.

%\newpage
\bibliography{main}
\bibliographystyle{main}

\newpage

\end{document}